%% file: main.tex
\documentclass{article}

\usepackage[round]{natbib}

\usepackage[preprint]{neurips_2025}

\usepackage[utf8]{inputenc} 
\usepackage[T1]{fontenc}

\input{macro_arxiv}
\usepackage[table]{xcolor}

\newcommand{\deltaevolve}{\mbox{$\mathop{\mathtt{DeltaEvolve}}\limits$}\xspace}

\newcommand{\semanticdelta}{\mbox{$\mathop{\mathtt{semantic\ delta}}\limits$}\xspace}

\title{DeltaEvolve: Accelerating Scientific Discovery through Momentum-Driven Evolution}

\begin{document}



\author{
\textbf{Jiachen Jiang}$^{1}$ \quad
\textbf{Tianyu Ding}$^{2}$ \quad
\textbf{Zhihui Zhu}$^{1}$.}

\affil{
$^{1}$Department of Computer Science and Engineering, The Ohio State University\\
$^{2}$Microsoft\\
\texttt{jiang.2880@osu.edu}, \quad
\texttt{tianyuding@microsoft.com}, \quad
\texttt{zhu.3440@osu.edu}

}

\maketitle

\begin{abstract}
LLM–driven evolutionary systems have shown promise for automated science discovery, yet existing approaches such as AlphaEvolve rely on full-code histories that are context-inefficient and potentially provide weak evolutionary guidance. In this work, we first formalize the evolutionary agents as a general Expectation–Maximization framework, where the language model samples candidate programs (E-step) and the system updates the control context based on evaluation feedback (M-step). Under this view, constructing context via full-code snapshots constitutes a suboptimal M-step, as redundant implement details dilutes core algorithmic ideas, making it difficult to provide clear inspirations for evolution. To address this, we propose \deltaevolve, a momentum-driven evolutionary framework that replaces full-code history with structured \semanticdelta capturing how and why  modifications between successive nodes affect performance. As programs are often decomposable, \semanticdelta usually contains many effective components which are transferable and more informative to drive improvement. By organizing \semanticdelta through multi-level database and progressive disclosure mechanism, input tokens are further reduced. Empirical evaluations on tasks across diverse scientific domains show that our framework can discover better solution with less token consumption over full-code-based evolutionary agents.
\end{abstract}

\input{sections/1_introduction}

\input{sections/2_related_works}
\input{sections/3_framework}

\input{sections/4_methods}

\input{sections/5_experiments}

\section{Acknowledgement}
We acknowledge support from NSF grants IIS-2312840 and IIS-2402952. We gratefully acknowledge Ismail Alkhouri, Yuxin Dong, Xia Ning, and Huan Sun for valuable discussions.

\bibliographystyle{unsrtnat}
\bibliography{reference}

\input{sections/appendix}

\end{document}

%% file: macro_arxiv.tex
\usepackage{amsmath,amssymb,amsbsy,amsfonts,amscd,bm,url,color,latexsym}
\usepackage{amsthm}
\usepackage[colorlinks=true,
            citecolor=blue,    
            linkcolor=red,    
            urlcolor=blue 
           ]{hyperref}

\usepackage{cleveref}  
\usepackage{paralist}
\usepackage[dvipsnames]{xcolor}
\usepackage{color}
\usepackage{graphicx}
\graphicspath{{./figs/}}
\usepackage{comment}
\usepackage{multirow}
\usepackage{enumitem}
\usepackage{fancyhdr}
\usepackage{cite}
\usepackage{subcaption}
\usepackage{authblk}
\usepackage{booktabs} 
\usepackage{wrapfig}
\usepackage{adjustbox}

\newtheorem{theorem}{Theorem}[section]

\newtheorem{definition}[theorem]{Definition}

\usepackage{enumitem}
\usepackage{tabularx}

\newcommand{\e}{\begin{equation}}
\newcommand{\ee}{\end{equation}}
\newcommand{\en}{\begin{equation*}}
\newcommand{\een}{\end{equation*}}
\newcommand{\eqn}{\begin{eqnarray}}
\newcommand{\eeqn}{\end{eqnarray}}
\newcommand{\bmat}{\begin{bmatrix}}
\newcommand{\emat}{\end{bmatrix}}

\DeclareMathAlphabet\mathbfcal{OMS}{cmsy}{b}{n}

\newcommand{\calA}{\mathcal{A}}

\newcommand{\calC}{\mathcal{C}}

\newcommand{\calP}{\mathcal{P}}

\graphicspath{{./figs/}}

\newlength{\imgwidth}
\usepackage{tikz}
\usepackage{tikz-3dplot}
\usepackage{pgfplots}
\usepgfplotslibrary{patchplots}
\usetikzlibrary{patterns, positioning, arrows}
\pgfplotsset{compat=1.15}

\usepackage[most]{tcolorbox}
\newtcolorbox{titlebox}[1]{%
    tikznode boxed title,
    enhanced,
    arc=0mm,
    boxrule=1pt,
    halign=left,
    bottom=1ex,
    interior style={white},
    attach boxed title to top left={xshift=3ex,yshift=-\tcboxedtitleheight/2},
    fonttitle=\bfseries,
    colbacktitle=white,coltitle=black,
    boxed title style={size=small,colframe=white,boxrule=0pt},
    title={#1}
}

\newtcolorbox{systempromptbox}{
  enhanced,
  breakable,
  colback=white,
  colframe=gray!110,
  boxrule=0.8pt,
  arc=2mm,
  left=6pt,
  right=6pt,
  top=6pt,
  bottom=6pt,
  fonttitle=\bfseries,
  fontupper=\ttfamily\small,
  coltitle=white,
  colbacktitle=gray!110,
  title=Symtem Prompt Template
}

\newtcolorbox{userpromptbox}{
  enhanced,
  breakable,
  colback=white,
  colframe=gray!110,
  boxrule=0.8pt,
  arc=2mm,
  left=6pt,
  right=6pt,
  top=6pt,
  bottom=6pt,
  fonttitle=\bfseries,
  fontupper=\ttfamily\small,
  coltitle=white,
  colbacktitle=gray!110,
  title=User Prompt Template
}

\newtcolorbox{level1box}{
  enhanced,
  breakable,
  colback=white,
  colframe=gray!110,
  boxrule=0.8pt,
  arc=2mm,
  left=6pt,
  right=6pt,
  top=6pt,
  bottom=6pt,
  fonttitle=\bfseries,
  fontupper=\ttfamily\small,
  coltitle=white,
  colbacktitle=gray!110,
  title=Delta Summary (Level 1)
}

\newtcolorbox{level2box}{
  enhanced,
  breakable,
  colback=white,
  colframe=gray!110,
  boxrule=0.8pt,
  arc=2mm,
  left=6pt,
  right=6pt,
  top=6pt,
  bottom=6pt,
  fonttitle=\bfseries,
  fontupper=\ttfamily\small,
  coltitle=white,
  colbacktitle=gray!110,
  title=Delta Plan Details (Level 2)
}

\usepackage{xspace}

\definecolor{codebg}{RGB}{235,245,255}

\lstdefinestyle{pyblue}{
  language=Python,
  basicstyle=\ttfamily\footnotesize,
  numbers=left,
  numberstyle=\tiny,
  stepnumber=1,
  numbersep=8pt,
  backgroundcolor=\color{codebg},
  frame=single,
  rulecolor=\color{black!20},
  breaklines=true,
  tabsize=4,
  showstringspaces=false,
  keywordstyle=\color{blue!70!black}\bfseries,
  commentstyle=\color{green!40!black},
  stringstyle=\color{orange!60!black}
}

\newcommand{\lblue}{\cellcolor[HTML]{EFF7FE}} 
\newcommand{\lorange}{\cellcolor[HTML]{FFF7E6}} 

%% file: sections/1_introduction.tex
\section{Introduction}

LLMs have facilitated automated science discovery across diverse domains such as mathematical optimization\citep{georgiev2025mathematical, hubert2025olympiad, trinh2024solving}, physical systems~\citep{li2025codepde}, and molecular discovery~\citep{averly2025liddia}. Although different tasks, the common core objective is to discover a high-performing object that satisfy desirable quantitative properties. Fundamentally, all these tasks can be represented as the search problem: solutions are difficult to synthesize directly but computationally easy to evaluate. Recent works~\citep{real2020automl, mankowitz2023faster} mainly choose to search in the code space, since code is sufficiently expressive to represent complex algorithms, aligns well with LLM pretraining, and enables automatic evaluation through execution. 

For simple problems, single-turn generation conditioned on problem specifications can already be effective. However, complex tasks typically require interaction with execution environments and iterative refinement based on feedback. This has led to self-evolving agents such as FunSearch~\citep{romera2024mathematical} and AlphaEvolve~\citep{novikov2025alphaevolve}, which iteratively generate programs, execute them, analyze feedback signals, and produce improved variants. These approaches can be interpreted as a form of test-time scaling, instead of relying solely on internal reasoning tokens, the performance improves by allocating more computation to accumulating history and feedback across iterations.

Despite their success, existing evolving agents face two fundamental limitations:
\begin{itemize}
[leftmargin=12pt,itemsep=2pt,topsep=0pt,parsep=0pt]
    \item \textbf{Limited context window.} Extensive multi-turn inference incurs prohibitive computational cost and latency, preventing effective utilization of the full evolution history~\citep{bansal2025let}. As a result, methods such as AlphaEvolve~\citep{novikov2025alphaevolve} retain only a small subset of high-quality solutions (e.g., top-performing or diverse programs). However, for complex systems, individual programs are already long, leaving limited capacity to reuse historical information.

    \item \textbf{Insufficient evolutionary guidance.} Complete programs often entangles core algorithmic ideas with extensive implementation and control logic which are irrelevant to the underlying strategy, making it hard for LLMs to isolate the truly useful components from vast code histories. Therefore, full codes would not explicitly capture transferable patterns of success or failure that could guide future iterations, causing missed successful patterns or repeated failures. 

\end{itemize}

This motivates our research question: \textit{how can we provide stronger evolutionary guidance while staying within a limited context budget?} To rigorously address this, we first formalize the evolutionary agents within a general Expectation–Maximization (EM) framework. Optimization proceeds by alternating two steps: in the E-step, the language model samples candidate programs conditioned on the current context; in the M-step, the system updates the control context using evaluation feedback to maximize the objective. From this perspective, the standard practice of representing history via static full-code snapshots constitutes a suboptimal M-step: while it captures the current state, it obscures which changes led to performance gains or losses, weakening the learning signal.

\begin{figure}[t]
    \centering
    \includegraphics[width=\linewidth]{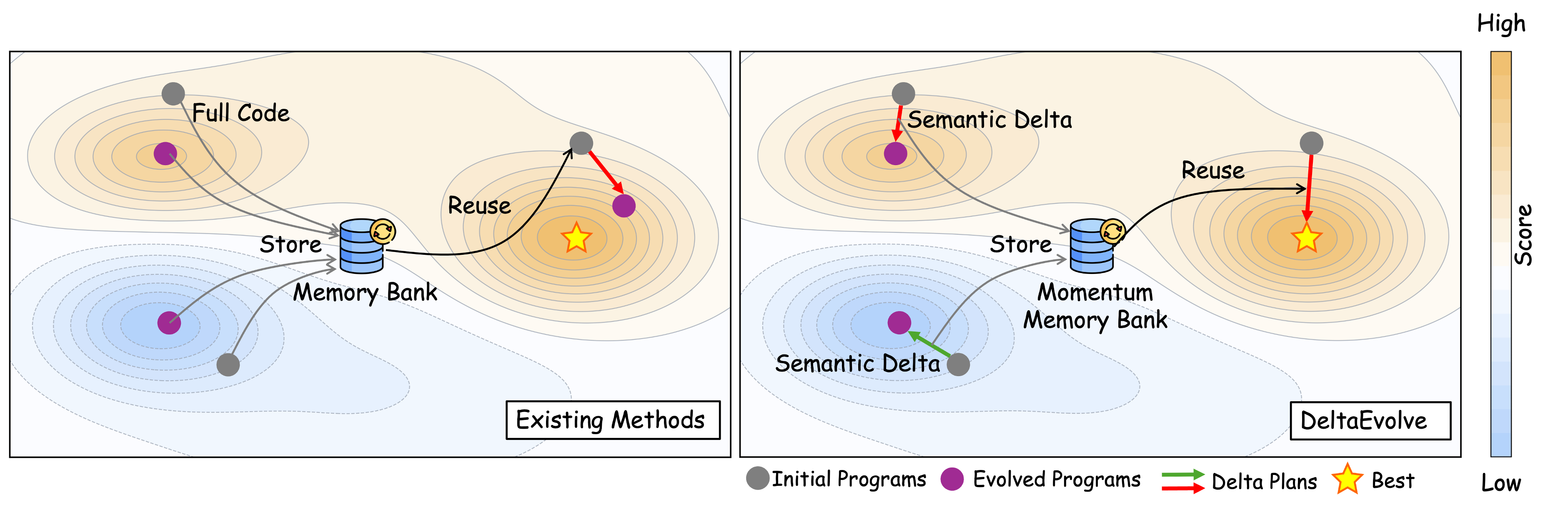}
    \caption{Illustration of search dynamics under existing methods and \deltaevolve. Existing methods store and reuse full programs, whereas \deltaevolve stores \semanticdelta that capture what changed and why it worked, forming a momentum-like memory that provides more informative guidance for reuse.}
    \label{fig:illu}
    \vspace{-.1in}
\end{figure}

We hypothesize that within a fixed task, the modifications that affect performance in earlier iterations often capture task-level structures and may remain informative for guiding updates to different programs in later iterations. These semantic changes are more informative for guiding future updates than the static solutions themselves. Based on this hypothesis, we propose \deltaevolve, a momentum-driven evolutionary framework designed to optimize the M-step. As illustrated in \Cref{fig:illu}, instead of storing full-code snapshots in the memory bank, we propose to use \semanticdelta to record the core semantic modifications from a parent node to its offspring and their resulting impact on performance. The memory bank stores \semanticdelta between pairs of nodes. As these deltas capture what changed across successive iterations and guide future updates, they form a directional signal analogous to the \textit{momentum} term in optimization.

To implement \semanticdelta efficiently, we augment each node with additional structures: a delta summary (Level-1) and delta plan details (Level-2), alongside the existing full code (Level 3), forming a pyramid-structured \texttt{Multi-Level Database}. Then the \texttt{Progressive Disclosure Sampler} dynamically decides the level of detail to expose for historical nodes based on relevance and recency: the current parent node is presented as full code for editing, while historical inspiration nodes are compressed into delta summaries or delta plan details. This approach exposes the logic of past improvements without wasting tokens on their implementation details.

\begin{figure}[t]
    \centering
    \includegraphics[width=0.5\linewidth]{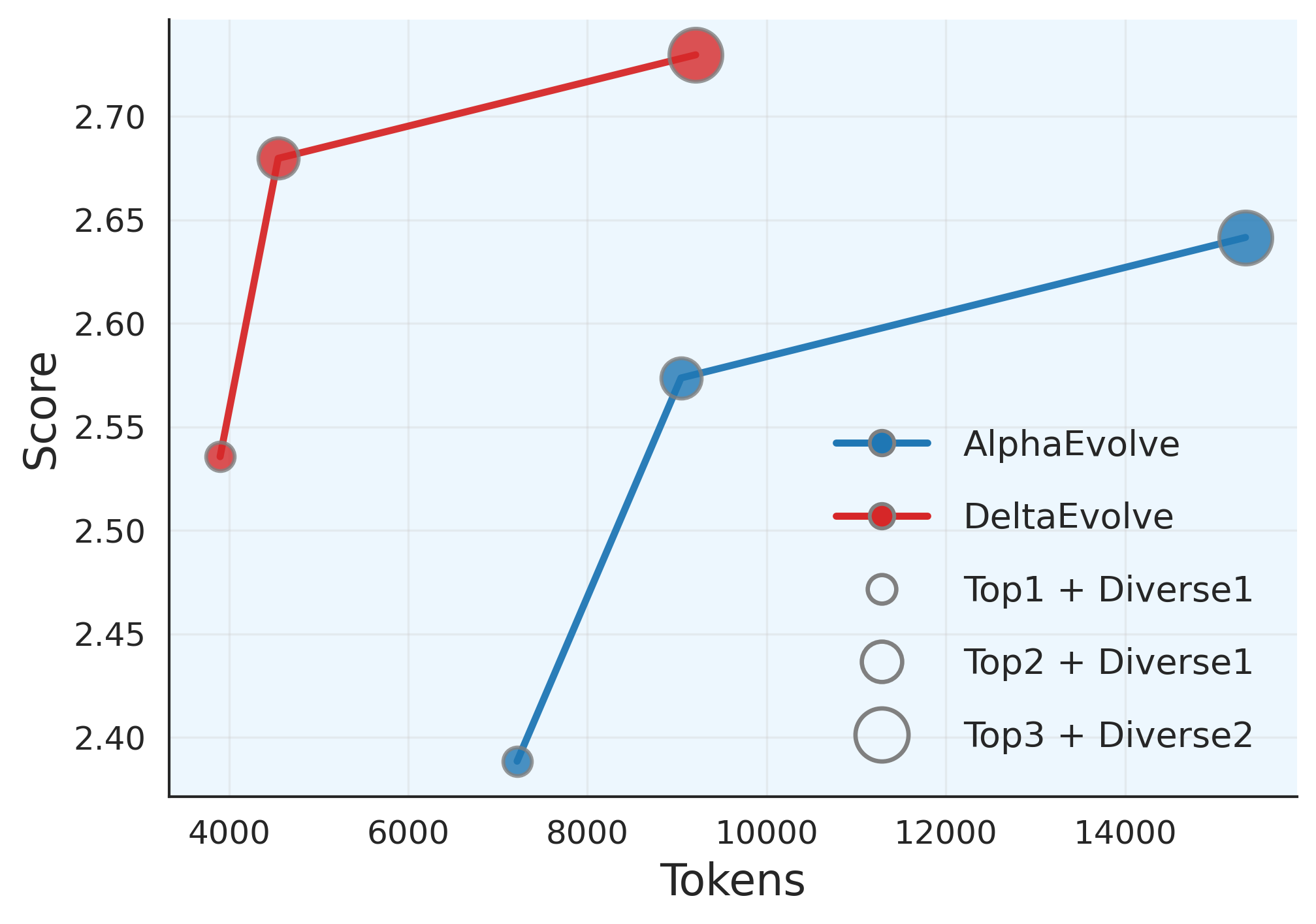}
    \caption{Comparison between AlphaEvolve and \deltaevolve (ours) on black-box optimization over 100 iterations. The x-axis shows cumulative consumed tokens across all iterations, and y-axis shows the best achieved score. Point size indicates the number of top and diverse nodes included in the context. \deltaevolve consistently achieves higher scores with fewer tokens. }
    \label{fig:compare}
    \vspace{-.2in}
\end{figure}

Our evaluation prioritize two critical metrics: solution quality and token consumption. As illustrated in \Cref{fig:compare}, on the black-box optimization task, \deltaevolve consistently achieves higher scores with significantly fewer input tokens across varying context hyperparameters (e.g., number of top and diverse programs). To ensure generality and robustness, we further validate our framework across five diverse domains. Extensive experiments confirm that \deltaevolve discovers solutions of comparable or superior quality to state-of-the-art baselines, all while reducing total token consumption by approximately 36.79\% on average.

Overall, our work makes the following contributions:
\begin{itemize}
[leftmargin=12pt,itemsep=2pt,topsep=0pt,parsep=0pt]

\item \textbf{General Evolutionary Framework.} We formalize the evolution as an Expectation–Maximization (EM) process, identifying updating the context (the M-Step) as the critical bottleneck in existing systems.

\item \textbf{DeltaEvolve.} We propose a momentum-driven framework that replaces static full-code snapshots with \semanticdelta. This approach utilizes a progressive disclosure mechanism to maximize context utility.

\item \textbf{Empirical Efficiency}: Through comprehensive evaluations across five scientific domains, we demonstrate that \deltaevolve achieves superior solution quality while reducing token consumption compared to state-of-the-art full-code baselines.
\end{itemize}

%% file: sections/2_related_works.tex
\section{Related Works}

\textbf{Evolutionary Coding Agents.}  The integration of LLMs with evolutionary algorithms has produced powerful frameworks like AlphaEvolve~\citep{novikov2025alphaevolve} and OpenEvolve~\citep{openevolve}, which established the core viability of LLM-driven program evolution. Subsequent works have significantly advanced LLM-driven evolution through improved orchestration. Building on this foundation, subsequent works have focused on optimizing the orchestration of the search process. Specifically, ShinkaEvolve~\citep{lange2025shinkaevolve}, GigaEvo~\citep{khrulkov2025gigaevo}, and CodeEvolve~\citep{assumpccao2025codeevolve} introduce advanced population management techniques, including novelty-based rejection sampling, MAP-Elites quality-diversity algorithms, and island-based crossover operators. Moreover, ThetaEvolve~\citep{wang2025thetaevolve} explores fine-tuning smaller LLMs via Reinforcement Learning. However, despite these unique optimization strategies, all these systems fundamentally rely on representing historical solutions as full source code within the context window. This approach imposes a rigid token bottleneck that forces a trade-off between history length and detail; in contrast, \deltaevolve resolves this by representing history as \semanticdelta, enabling the agent to evolve with minimal token cost.

\textbf{Context Engineering.} Recent work on agent prompting and context engineering has investigated how to structure and compress context to support long-horizon reasoning and iterative agent behavior. Practical agent systems and guidelines~\citep{anthropic2025context} emphasize modular prompts and progressive disclosure, while memory-augmented agents such as MemoryLLM ~\citep{wang2024memoryllm} and its scalable extension M+~\citep{wang2025m+} introduce long-term memory to mitigate context overflow in extended interactions. In parallel, prompt and context compression methods~\citep{zhang2024long, fei2025efficient, shi2025no} aim to reduce effective context length through token selection or adaptive refinement while preserving task-relevant information. However, these existing techniques are inherently generic; they compress information without awareness of optimization objectives. In contrast, our work does not merely reduce context length, but actively constructs context that directs the agent toward higher-quality solutions.

%% file: sections/3_framework.tex
\section{Evolutionary Framework}
\label{sec:evolutionary_framework}

In this section, we aim to examine LLM-driven evolution from a rigorous optimization perspective.

\subsection{Problem Definition}

We consider a broad class of open-ended problems arising in mathematics, scientific discovery, and engineering. These tasks include, but are not limited to, classical optimization problems, algorithm and code design, combinatorial constructions, heuristic discovery, and scientific or engineering workflows where the solution structure or algorithmic strategy is not known a priori and must be discovered through interaction with an evaluator, rather than optimized within a fixed parameterization.

In this work, we focus on problems whose solutions, or the procedures for discovering solutions, can be expressed as programs. Programs provide a flexible and executable representation that supports structured modification, composition, and verification, making them particularly well suited for iterative improvement. But we note that the same principles apply to other forms of solution representations (such as symbolic constructions or mathematical proofs) provided they admit an evaluator that offers task-dependent feedback. Programs serve here as a concrete instantiation that enables precise execution and scalable evaluation.

Formally, let $\calP$ denote the space of programs. For a given task or problem $q$, let $R_q: \calP \rightarrow \mathbb{R}$
be an evaluator that assigns a scalar score or reward (e.g., accuracy, runtime, objective value, or combinations) to each program by executing it. The goal is to find a program $p^\star$ that acheives the highest possible score, i.e., 
\begin{equation}
    p^* = \arg\max_{p \in \mathcal{P}} R_q(p),
\label{eq:obj}\end{equation}
Although \eqref{eq:obj} resembles a conventional optimization objective, it differs from classical mathematical optimization in several fundamental ways: $(i)$ the task specification $q$ may be partially formal and partially expressed in natural language, making the objective implicitly defined through the evaluator rather than an explicit analytic form, $(ii)$ the search space $\calP$ consists of programs with variable length, control flow, and data structures, resulting in a discrete, extremely large space without a fixed dimensionality or natural notion of locality as in $\mathbb{R}^d$ space, $(iii)$  the evaluator $R_q$ is procedural and non-differentiable, often involving program execution and hard correctness constraints, which together induce a highly irregular and discontinuous objective landscape, and $(iv)$ candidate programs may themselves encode search or optimization procedures, so the problem amounts to optimizing processes that perform optimization internally. These characteristics place \eqref{eq:obj} outside the scope of classical optimization methods and motivate search-based, feedback-driven approaches such as AlphaEvolve. Example in~\Cref{tab:concrete_forms} further illustrates the difference.

\subsection{The EM Interpretation}

Since the objective function\footnote{To simplify the notation, when it is clear from the context, we will simply denote the evaluator $R_q$ via $R$.} $R$ is the only accessible feedback mechanisms (analytical gradients $\nabla R$ are inaccessible), the discovery task falls under the domain of zero-order black-box optimization. A standard approach for such problems is to employ an iterative Expectation-Maximization (EM) strategy: first, sampling from a distribution model(e.g. multivariate normal distribution)~\citep{hansen2016cma} with latent variables (E-Step), and subsequently updating the latent variables based on the observed evaluation (M-Step).

However, the reward landscape in scientific discovery is exceptionally sparse and complex, rendering traditional surrogate models insufficient. Therefore, current agents increasingly leverage the reasoning capabilities of LLMs as a powerful distribution model. Its weight parameters and context would serve as the latent variables. We use $\calA_{\theta}(q\oplus \calC)$ to denote an LLM that returns a likely program $p$ given the problem description $q$ and additional context $\calC$, where $q \oplus \mathcal{C}$ denotes the concatenation of the two text inputs. In this paper, to accommodate the use of frontier LLMs, we assume that the model parameters $\theta$ are fixed and that only black-box access via an LLM API is available. We therefore treat the context $\calC$ as a latent variable to be optimized. Under this formulation, we solve the objective in \eqref{eq:obj} via
\begin{align}
\max_{\calC} R(\calA_{\theta}(q\oplus \calC)).
\label{eq:newobj}
\end{align}
Although it appears similar to prompt optimization or engineering \citep{ramnath2025systematic}, \eqref{eq:newobj} differs fundamentally in that it focuses on {\it problem-specific context optimization}. Rather than learning a generic prompt that improves average performance across tasks, the context $\mathcal{C}$ is tailored to a given problem instance $q$ and optimized to maximize the objective $R$, acting as a {\it latent}, adaptive variable rather than a reusable instruction template. Intuitively, conditioning on $\mathcal{C}=p^\star$ increases the likelihood of generating $p^\star$ or related high-quality programs, whereas poor or random contexts degrade performance. {\it From a Bayesian perspective, $\mathcal{C}$ functions as an inductive bias over programs, with more informative contexts shaping the conditional distribution of the LLM toward higher-reward solutions.} 
However, directly optimizing $\mathcal{C}$ is challenging due to the high-dimensional, discrete, and non-convex objective. We therefore adopt an iterative optimization strategy inspired by the expectation–maximization (EM) framework, which alternates between refining the distribution over candidate programs and updating context to increase expected reward.

\begin{enumerate}[leftmargin=12pt,itemsep=4pt,topsep=4pt,parsep=0pt]
    \item \textbf{E-Step (Sampling).} 
Given the current context $\mathcal{C}_t$, the system samples candidate programs to estimate the local landscape. We also query the evaluator to obtain the history $\mathcal{H}_{\text{new}}$:
\begin{equation}
\mathcal{H}_{\text{new}} = \left\{ (p_t, R(p_t)) \mid p_t = \calA_{\theta}(q\oplus \calC) \right\}.
\end{equation}

 \item \textbf{M-Step (Context Update).} 
The system updates the context to $\mathcal{C}_{t+1}$ to maximize the expected score in the next iteration. Using the accumulated history $\mathcal{H}_{t+1} = \mathcal{H}_{t} \cup \mathcal{H}_{\text{new}}$, also referred to as the memory bank, we define the update via a policy $\pi$:
\begin{equation}
    \mathcal{C}_{t+1} = \pi(\mathcal{H}_{t+1}).
\end{equation}

It reveals that the policy of constructing context from history plays a key role in updating the search distribution and guiding evolution.

\end{enumerate}

This framework provides a unified perspective on system improvement. In this setting, agents typically assume fixed model weights\footnote{Updating $\theta$ corresponds to E-Step optimization, which has been explored in very recent works like ThetaEvolve~\citep{wang2025thetaevolve} and TTT-Discover~\citep{yuksekgonul2026learning}.}, thereby focusing the optimization on the M-Step: designing the policy $\pi$ to construct the most effective context from history.  A naive strategy, such as greedy refinement, builds the context using a single program, such as the most recent or the highest-performing one. Conditioned on this program, the LLM attempts to modify it to better solve the target problem; we refer to this program as the parent program ${p_{\text{parent}}}$.
This procedure is analogous to classical hill-climbing in descent-based optimization methods such as gradient descent. However, when conditioned solely on the parent program, the agent is susceptible to {\it mode collapse}, in which it repeatedly produces similar modifications across iterations, thereby limiting exploration and hindering escape from local optima.

Evolutionary system like AlphaEvolve~\citep{novikov2025alphaevolve} attempt to enhance this by including $n$ ``inspiration nodes" in the context as,
\begin{equation}
    \mathcal{C}_{t+1} = \{ p_{\text{parent}} \oplus \underbrace{\{(p_1, R_1),\dots, (p_n, R_n)}_{\text{Full Code Inspirations}} \}\}.
    \label{eq:alphaevolve_context}
\end{equation}
Each node stores the full program together with its evaluation score returned by the evaluator. The set of inspiration nodes typically consists of Top-$k$ and Diverse-$m$ programs. Top-$k$ nodes are selected based on the evaluator feedback $R$, while Diverse-$m$ nodes are chosen according to the semantic similarity of code texts. These inspiration programs provide additional contextual signals that guide program modification, analogous to leveraging neighboring points to estimate a descent direction in gradient-based optimization.

\textbf{Significance of Framework.} Evolving systems are traditionally viewed through the lens of agentic interaction—similar to reinforcement learning (RL)—where an agent maximizes reward through feedback. However, unlike RL, these systems cannot perform parameter updates, leaving the mechanisms of convergence and ``learning" theoretically ambiguous. By formalizing evolution as a black-box optimization problem within an EM framework, our framework isolates the true driver of the system. It reveals that in the absence of weight updates, the context $\mathcal{C}$ acts as the sole proxy for learned variables. 
Consequently, the M-step plays a role analogous to a gradient update, a connection we exploit in the next section to develop more efficient optimization strategies.

\subsection{Context Selection Dominates Scalar Feedback}
\label{sec:analysis_guidance}

While evaluator feedback $R$ serves as the directional signal for evolution, the precise mechanism driving the optimization remains unclear. Our framework reveals that the objective $R$ serves a dual role in the M-Step: 

\begin{enumerate}[leftmargin=12pt,itemsep=4pt,topsep=4pt,parsep=0pt]
    \item \textbf{Context Selection.} Guiding selection policy $\pi$ to choose high-quality codes from memory bank into context window $\mathcal{C}$ (e.g., filtering high-quality solutions); 
    
    \item \textbf{Scalar Condition.} Providing numerical feedback correlates to the code directly within $\mathcal{C}$ (e.g., ``Score: 0.95") to condition the LLM's next generation. 

\end{enumerate}
    
This duality raises a fundamental question regarding the nature of LLM-based optimization: \textit{Does the performance gain arise from mimicking the high-quality solutions preserved by the selection policy, or from correlating code with numerical scores?} To answer this question, we decouple their effects and design the following controlled experiments.

\textbf{Experimental Design.} We conduct a controlled study on problems from five domains using the AlphaEvolve framework. We compare three distinct settings:
\begin{itemize}
[leftmargin=12pt,itemsep=4pt,topsep=4pt,parsep=0pt]
    \item \textbf{AlphaEvolve (Standard):} Uses Top-$k$ selection ($\pi_{\text{elite}}$) and includes explicit scores in the prompt.
    \item \textbf{Blind-Elite:} Uses Top-$k$ selection but \textbf{masks} all numerical scores from the prompt. The LLM receives the best codes but not the values.
    \item \textbf{Random-Context:} Randomly select the historical programs ($\pi_{\text{rand}}$) from the histories but including scores.
\end{itemize}

\textbf{Results \& Insight.} As shown in Table~\ref{tab:ablation_raw}, the results are revealing.
Removing numerical scores (\textit{Blind-Elite}) results in similar performance compared to the AlphaEvolve. In contrast, removing the selection policy (\textit{Random-Context}) causes performance to collapse, even when scores are clearly visible.

\begin{table}[h]
\centering
\caption{Ablation study on evaluator feedback mechanism (using \texttt{gpt-5-mini/o3-mini}). We compare \texttt{AlphaEvolve} (Standard), \texttt{Blind-Elite} (No Score), and \texttt{Random-Context} (No Selection) using objective scores.}
\label{tab:ablation_raw}
\begin{tabular}{@{}l | c c c@{}}
\toprule
\multirow{2}{*}{\textbf{Task}} & \textbf{AlphaEvolve} & \textbf{Blind-Elite} & \textbf{Random-Context} \\
& \small{(Standard)} & \small{(No Score)} & \small{(No Selection)} \\
\midrule
1. Blackbox Optimization & \textbf{2.642} & 2.578 & $1.429$ \\
2. Hexagon Packing & \textbf{0.972} & 0.970 & $0.786$ \\
3. Symbolic Regression & \textbf{3.265} & $3.179$ & $2.576$ \\
4. PDE Solver  & \textbf{0.885} & 0.884 & $0.803$ \\
5. Efficient Convolution & $0.897$ & \textbf{0.911} & $0.550$ \\
\bottomrule
\end{tabular}

\end{table}

\textbf{Implication} These findings suggest that standard evolutionary agents operate primarily via in-context learning, not implicit regression. The agent improves by utilizing patterns from the high-quality context selected by the evaluator. The scalar feedback fails to guide the LLM because it provides a magnitude of success without explaining the mechanism of improvement.

%% file: sections/4_methods.tex
\section{DeltaEvolve}
\label{sec:deltaevolve}

In this section, we first motivate replacing full-code with \semanticdelta (\Cref{sec:motivation}), then describe their implementation via a multi-level database (\Cref{sec:multilevel-db}) and a progressive disclosure sampler (\Cref{sec:progressive}).

\subsection{Motivation}
\label{sec:motivation}

Building on the framework in~\Cref{sec:evolutionary_framework}, we identify the \texttt{M-Step} as the decisive factor in agent design, as it governs the evolutionary trajectory.  The core challenge lies in designing more effective strategy $\pi$ to construct the context $\mathcal{C}$ from history that effectively guides the search.

Here, we argue implementing \texttt{M-step} using \eqref{eq:alphaevolve_context} is still suboptimal: it populates $\mathcal{C}$ with full code snapshots and raw numerical scores. It provides a static \textit{state description} rather than an \textit{update gradient}. While it captures where the search is, it obscures the trajectory of how it got there. To address this, we first observe that in the domain of algorithmic discovery, programs are not monolithic entities; they are \textit{compositional programs}.
\begin{definition}[\textbf{Compositional Programs}] A program $p$ is defined as a structured composition $p = \Phi(\ell_1, \ell_2, \dots, \ell_J)$ ~\citep{ellis2023dreamcoder}, where each $\ell_j$ represents a reusable functional component (e.g., subroutines, API calls, or functional blocks) and $\Phi$ denotes the composition operator. 
\label{def:compositional}
\end{definition}
LLMs drive evolution through \textit{compositional generalization}—the capacity to decompose complex tasks and recombine existing components into novel structures \citep{reed2015neural, shi2022compositional}. Based on this insight, the drivers of evolution are not entire solutions, but some effective components. These logical modifications are often transferable and more informative than the static solutions themselves. Therefore, we introduce \semanticdelta to explicitly capture these transferable components:
\begin{equation}
\delta_i = \mathrm{Diff}(p_i, p_{i-1}).
\end{equation}
Here, $\delta_i$ represents the interpretable logic change between a program and its parent. The $\delta_i$ usually consists of multiple isolated modification components, and each component records the exact logic of which part is changed in concise natural-language description. Refer to \Cref{fig:level2_example} in the appendix for an example of these decomposed components.

\begin{figure}[t]
    \centering\includegraphics[width=0.7\linewidth]{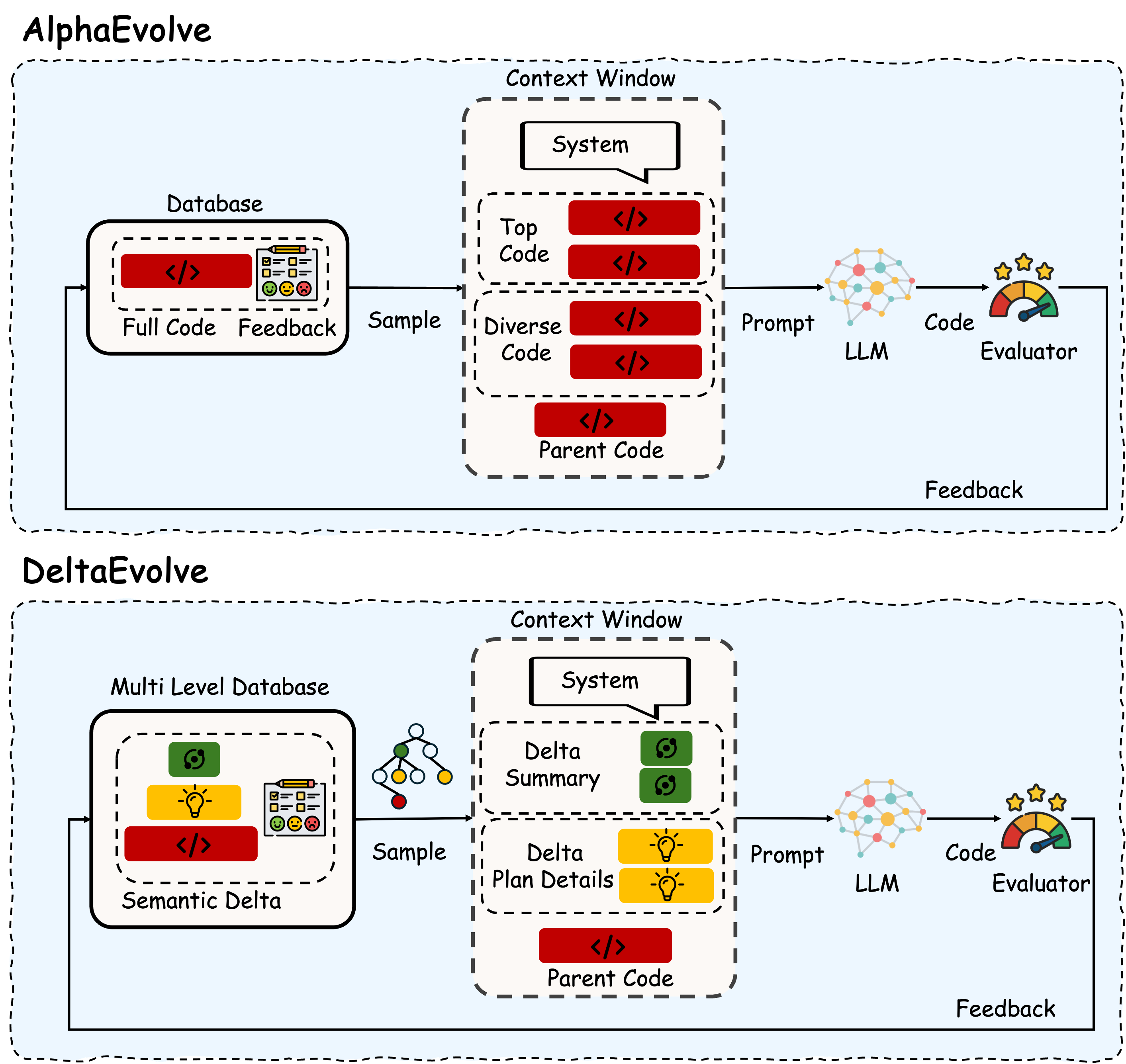}
    \caption{Comparison of the \deltaevolve pipeline with AlphaEvolve. \deltaevolve incorporates \semanticdelta into the context window instead of full code.  }
    \label{fig:main}
    \vspace{-.2in}
\end{figure}

By accumulating \semanticdelta across different nodes, we construct the context of \deltaevolve as,
\begin{align}
    \mathcal{C}_{\text{delta}} = \{ p_{\text{parent}} \oplus \underbrace{(\delta_{1}, \Delta R_{1}), \dots, (\delta_{n}, \Delta R_{n})}_{\text{Delta Inspirations}} \}
\end{align}
where $\Delta R$ denotes a qualitative performance shift (e.g., ``Improved or Degraded") rather than a precise numerical change (e.g., ``$\pm 0.15$"). As analyzed in Section~\ref{sec:analysis_guidance}, exact numerical gains are often less informative and unnecessary.

This formulation closely aligns with the concept of momentum in classical optimization. Just as SGD methods accumulate gradients through successive differences between iterates ($x_i - x_{i-1}$), the \semanticdelta capture the differences between successive programs ($p_i, p_{i-1}$), In this sense, the accumulated deltas serve as a discrete, semantic analogue of a momentum vector, encoding the prevailing direction of improvement across iterations.

By eliminating redundant implementation details, this representation substantially improves token efficiency, enabling the model to attend to a longer and more informative history of algorithmic evolution. To further mitigate token overhead in practice, we show the implementation details of \semanticdelta in the following sections.

\subsection{Multi-Level Database}
\label{sec:multilevel-db}

To implement \semanticdelta efficiently, we augment each node with two additional structures: a delta summary (Level-1) and delta plan details (Level-2). Together with the existing full code (Level 3), each node in the database is represented as a pyramid structure:
\begin{align}
N_t = (\delta^{(1)}_t,\; \delta^{(2)}_t,\; p_t),
\end{align}
where both $\delta^{(1)}_t$ and $\delta^{(2)}_t$ serve as the \semanticdelta representations—distinguished primarily by their token numbers—while $p_t$ represents the full executable code.
\paragraph{L1: Delta Summary ($\delta^{(1)}_t$).}
$\delta^{(1)}_t$ describes the overall strategy change from the parent node.
It focuses on high-level ideas and avoids implementation details, providing a
compact description of the search direction (typically 20--40 tokens).
We use an explicit \texttt{FROM/TO} format:
\begin{equation}
\delta^{(1)}_t := \langle \texttt{FROM: } s_{t-1} \quad \texttt{TO: } s_{t} \rangle,
\end{equation}
where $s_{t}$ is the high-level idea summary of the node $t$. Please refer to \Cref{fig:level1_example} for an example.

\paragraph{L2: Delta Plan Details ($\delta^{(2)}_t$).}
$\delta^{(2)}_t$ is a structured plan consisting of $J$ multiple modifications.
For each modification, we contrast the old logic with the new logic and state the
underlying hypothesis. This representation makes the reason for improvement
explicit, allowing the agent to learn what to change and why without reading the full code:
\begin{equation}
\delta^{(2)}_t := \{(\ell_{t-1}^{(j)}, \ell_{t}^{(j)}, \texttt{hyp}_t^{(j)})\}_{j=1}^J,
\end{equation}
where $\ell_{t}^{(j)}$ denotes the $j$-th logic component of node $t$, and
$\texttt{hyp}_{t}^{(j)}$ is the hypothesis predicted by the LLM to explain the modification. Please refer to \Cref{fig:level2_example} for an example.

\paragraph{L3: Full Code ($p_t$).} $p_t$  stores the full program of node $N_t$ and remains essential for evolution,
since node evaluation relies on executable code and new nodes must be derived by modifying the parent program rather than abstract plans.

All three levels are co-generated by the LLM during mutation. We enforce a strict output
format requirements (as shown in \Cref{fig:system_prompt}) that requires LLM generate all three levels in order. A lightweight parser extracts content between explicit delimiters (e.g., \texttt{\#Delta-Summary-Start} and \texttt{\#Delta-Summary-End}), enabling reliable retrieval of different parts.

\subsection{Progressive Disclosure Sampler}
\label{sec:progressive}

To further reduce token consumption, we employ a progressive disclosure mechanism
to control the level of detail exposed in context. Rather than treating all
historical nodes uniformly, the prompt sampler adaptively adjusts the abstraction
level of each node based on its relevance and recency, revealing either
coarse-grained summaries or finer-grained deltas as needed to guide the next mutation.

\begin{figure}[h!]
    \centering
    \includegraphics[width=0.5\linewidth]{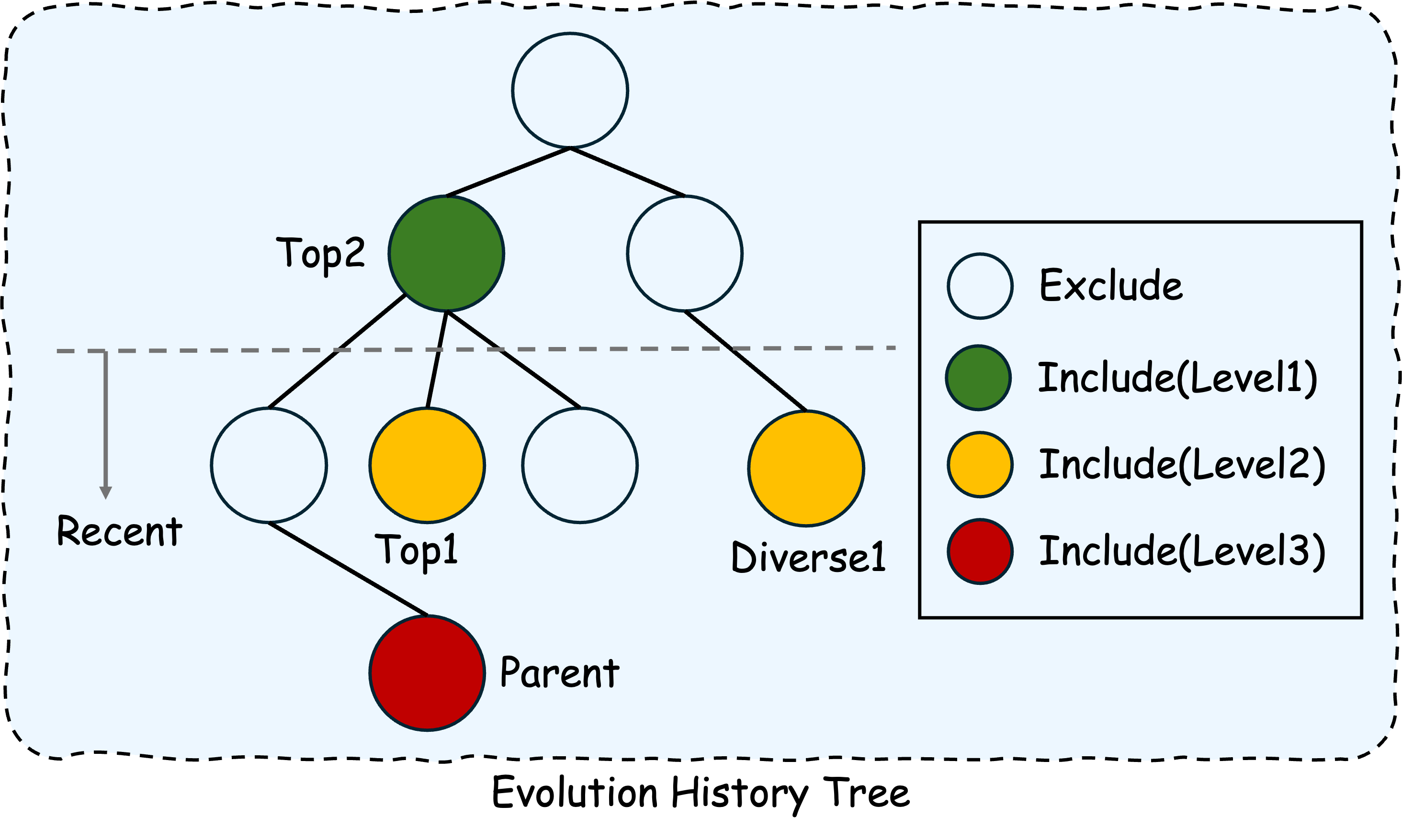}
    \caption{Progressive Disclosure Sampler}
    \label{fig:sampler}
    \vspace{-.1in}
\end{figure}

The sampler operates in two stages:
(1) \texttt{Node Selection}, which selects historical nodes for inclusion, and (2) \texttt{Multi-Level Rendering}, which assigns an appropriate abstraction level (Level~1,~2, or~3) to each selected node.
\paragraph{Phase 1: Node Selection.}
Given the database $\mathcal{D}$ at step $t$, we select three distinct sets of nodes
to construct the context:

\begin{enumerate}[leftmargin=12pt,itemsep=2pt,topsep=0pt,parsep=0pt]
    \item \textbf{Parent Node} ($N_{\text{parent}}$):
    The solution selected for modification. The parent is sampled using a stochastic
    selection policy $\pi_{\text{select}}$ that prioritizes high-reward nodes while
    retaining a non-zero probability of selecting lower-reward ones, thereby
    balancing exploitation and exploration.

    \item \textbf{Elite Nodes} ($\mathcal{N}_{\text{top}}$):
    A set of the $k$ high-scoring solutions discovered so far. These
    nodes represent successful strategies and capture the dominant directions of optimization.

    \item \textbf{Diverse Nodes} ($\mathcal{N}_{\text{div}}$):
To mitigate mode collapse, we sample $m$ diverse nodes based on the similarity of
their text-embedding representations. Specifically, nodes are organized into a
MAP-Elites grid\footnote{MAP-Elites (Multi-dimensional Archive of Phenotypic Elites)
is a quality-diversity algorithm that partitions the solution space into cells
based on behavior descriptors and retains the highest-performing solution within
each cell.}, and candidates are selected from cells that are maximally distant
from the parent, thereby injecting novel logic into the prompt.
\end{enumerate}

\paragraph{Phase 2: Progressive Rendering.}
After selecting the relevant nodes, we render them into the context with different levels of detail:

\begin{itemize}[leftmargin=12pt,itemsep=2pt,topsep=0pt,parsep=0pt]
    \item \textbf{Level 1 (Ancient History).}
For older elite nodes, we include only a concise delta summary $\delta^{(1)}$.
This provides a lightweight view of past strategy shifts without exposing
implementation details.

\item \textbf{Level 2 (Recent Insights).}
For more recent nodes and all selected elite or diverse inspirations,
we include the delta plan details $\delta^{(2)}$, which describe the concrete logic
changes and underlying hypotheses that proved effective.

\item \textbf{Level 3 (Immediate Context).}
For the current parent node, we include the full executable code
$p_{\text{parent}}$ to support
direct and valid code modification.
\end{itemize}

This progressive disclosure mechanism preserves informative evolutionary signals
while further reducing token usage, with the detailed algorithm described in~\Cref{alg:multilevel_db}.

\begin{table*}[t]
\label{tab:main_exp}
\centering
\caption{Comprehensive evaluation on tasks across 5 domains. We compare \deltaevolve against four distinct baselines: (1) \texttt{Parallel Sampling} (Best-of-N sampling), (2) \texttt{Greedy Refine} (Refinement top code), and (3) \texttt{AlphaEvolve} (Evolutionary search with full code). \texttt{Best Score} reports the objective value. \texttt{Token Consump.} represents the number of total used tokens.}
\label{tab:full_comparison}
\resizebox{\textwidth}{!}{
\begin{tabular}{@{}l | l | cc | cc@{}}
\toprule
\multirow{2.5}{*}{\textbf{Task Domain}} & \multirow{2.5}{*}{\textbf{Method}} & \multicolumn{2}{c|}{\textbf{GPT-5-mini + o3-mini}} & \multicolumn{2}{c}{\textbf{Gemini-2.5-flash-lite + Gemini-2.5-flash}} \\
\cmidrule(l){3-6}
& & \textbf{Best Score}($\uparrow$) & \textbf{Token Consump.}($\downarrow$) & \textbf{Best Score}($\uparrow$) & \textbf{Token Consump.}($\downarrow$) \\
\midrule
\multirow{4}{*}{\textbf{1. BlackBox Optimization}} 
&  Parallel Sampling & $0.4161$ & 390872 & $0.4161$ & 390872 \\
& Greedy Refine & $2.2618$ & 555430 & $2.3403$ & 1054566 \\
& \lblue AlphaEvolve (Full Code) & \lblue \underline{2.6415} & \lblue 1852841 & \lblue \underline{2.5221} & \lblue 1894890 \\
& \lorange \textbf{DeltaEvolve (Ours)} & \lorange \textbf{2.7297} & \lorange 1390709 & \lorange \textbf{3.9372} & \lorange 1227388 \\
\midrule
\multirow{4}{*}{\textbf{2. Hexagon Packing}} 
& Parallel Sampling & $0.4913$ & 565016 & $0.4913$ & 565016 \\
& Greedy Refine & $0.8508$ & 717697 & $0.4913$ & 814092 \\
& \lblue AlphaEvolve (Full Code) & \lblue \underline{0.9721} & \lblue 905249 &  \lblue \underline{0.7859} & \lblue 1334339 \\
& \lorange  \textbf{DeltaEvolve (Ours)} & \lorange \textbf{0.9821} & \lorange 827884 & \lorange \textbf{0.8804} & \lorange 893749 \\
\midrule
\multirow{4}{*}{\textbf{3. Symbolic Regression}} 
& Parallel Sampling & $1.8527$ & 503098 & $1.8535$ & 503098 \\
& Greedy Refine & $2.9553$ & 811404 & $2.9550$ & 756762 \\
& \lblue AlphaEvolve (Full Code) & \lblue \underline{3.2657} & \lblue 1660699 & \lblue \underline{3.2174} &  \lblue 1545091 \\
& \lorange \textbf{DeltaEvolve (Ours)} & \lorange \textbf{3.4174} & \lorange 810354 & \lorange \textbf{3.2198} & \lorange 832164 \\
\midrule
\multirow{4}{*}{\textbf{4. PDE Solver}} 
& Parallel Sampling & $0.7506$ & 154016 & $0.7506$ & 154016 \\
& Greedy Refine & $0.7506$ & 375186 & $0.7506$ &  255332 \\
& \lblue AlphaEvolve (Full Code) & \lblue \underline{0.8850} & \lblue 711298 & \lblue \underline{0.9901} & \lblue 595094 \\
&  \lorange \textbf{DeltaEvolve (Ours)} & \lorange \textbf{0.8915} & \lorange 562848 & \lorange \textbf{0.9931} & \lorange 253719 \\
\midrule
\multirow{4}{*}{\textbf{5. Efficient Convolution}} 
& Parallel Sampling & $0.8035$ & 139958 & $0.8036$ & 139958 \\
& Greedy Refine & $0.8035$ & 240473 & $0.8037$ & 305334 \\
& \lblue AlphaEvolve (Full Code) & \lblue \underline{0.8974} & \lblue 683439 & \lblue \underline{0.8219} & \lblue 885592\\
& \lorange \textbf{DeltaEvolve (Ours)} & \lorange \textbf{0.9067} & \lorange 348539 & \lorange \textbf{0.9032} & \lorange 517281 \\
\bottomrule
\end{tabular}
}
\vspace{-.2in}
\end{table*}

%% file: sections/5_experiments.tex
\section{Experiments}

In this section, we conduct experiments to answer the following questions:
\begin{itemize}
[leftmargin=12pt,itemsep=2pt,topsep=0pt,parsep=0pt]
    \item \textbf{RQ1 (Solution Quality):} \textit{Can \deltaevolve discover solutions with superior objective scores compared to state-of-the-art evolutionary baselines across diverse scientific domains?}
    \item \textbf{RQ2 (Token Consumption):} \textit{Does \deltaevolve reduce the computational cost (input tokens) required to discovery high-quality solutions?}
\end{itemize}
\subsection{Setup}

\paragraph{Tasks.} We evaluate open-ended problems across diverse domains: (1) \texttt{BlackboxOptimization}~\citep{bbob2019}: minimize the objective value of five standard black-box functions (e.g., Rosenbrock, Rastrigin) with input dimensions scaling from 3 to 40; (2) \texttt{HexagonPacking}~\citep{georgiev2025mathematical}: pack $N=11$ unit regular hexagons into the smallest possible outer regular hexagon without overlap; (3) \texttt{SymbolicRegression}~\citep{shojaee2025llm}: discover the mathematical expression that best fits a given dataset by uncovering latent relationships; (4) \texttt{PDESolver}~\citep{li2025codepde}: evolve numerically stable iterative linear solvers to minimize the residual norm of $Ax=b$ for large sparse systems; and (5) \texttt{EfficienctConvolution}~\citep{press2025algotune}: minimize the wall-clock execution time of 2D convolution kernels on dynamic problem scales ($30n \times 30n$) subject to strict correctness verification.
\paragraph{Baselines.} We compare \texttt{DeltaEvolve} against other code-generation paradigms: 
 (1) \texttt{Parallel Sampling}: a Best-of-$N$ sampling strategy that selects the highest-scoring candidate from independent samples; (2) \texttt{Greedy Refine}: iterative modification of the current best solution based on verbal feedback from the evaluator; and (3) \texttt{AlphaEvolve}~\citep{novikov2025alphaevolve}: a representative state-of-the-art evolutionary agent that relies on full-code history; we reproduce it using the open-source implementation of \texttt{OpenEvolve} ~\citep{openevolve}.
\paragraph{Implementations.} Following AlphaEvolve~\citep{novikov2025alphaevolve}, we employ an LLM ensemble strategy with probability of 0.8 for high-throughput generation, and 0.2 for complex reasoning across two distinct model families: (1) \texttt{gpt-5-mini} paired with \texttt{o3-mini}, and (2) \texttt{gemini-2.5-flash-lite} paired with \texttt{gemini-2.5-flash}. For the \texttt{Progressive Disclosure Sampler}, we configure the prompt context to include $k=3$ top-performing nodes and $m=2$ diverse nodes. Detailed configurations are provided in \Cref{tab:system_config} in the Appendix.

\subsection{Evaluation Metrics}
\label{sec:metrics}

\textbf{1. Best Score ($\uparrow$).} We report the raw objective value of the best-performing program discovered within the fixed budget, serving as the direct measure of solution quality.

\textbf{2. Token Consumption ($\downarrow$).} We measure the computational cost of discovery by the cumulative number of tokens consumed throughout the search process. Token consumption is defined as the total number of tokens used across all language model calls and all iterations, including parallel samples. Lower token consumption indicates lower overall computational cost.

\subsection{Main Results} \Cref{tab:main_exp} reports results across all task domains. To ensure robustness, we run each method with three random seeds ($11, 42, 100$). We report the maximum \texttt{best score} to capture best solutions achieved and the average \texttt{token consumption} to measure cost-effectiveness. \deltaevolve consistently outperforms all baselines on both metrics. Although gains in \texttt{best score} may appear small for some tasks, improvements near theoretical or best-known optima are increasingly difficult, making even modest gains meaningful. In terms of cost, \deltaevolve achieves substantially lower \texttt{token consumption} than AlphaEvolve, validating our core design: replacing redundant full-code snapshots with compact \semanticdelta representations reduce token consumption, enabling faster and more directed evolution.
\vspace{-.1in}
\section{Conclusion}
We present \deltaevolve, a novel momentum-driven framework that addresses the context inefficiency and weak evolutionary signals of existing full-code self-evolving systems. By formalizing program evolution as an Expectation-Maximization process, we introduce \semanticdelta as a transferable and informative driver of improvement. Together, the multi-level database and the progressive disclosure sampler enable a more efficient use of context. Experiments demonstrate that \deltaevolve consistently outperforms state-of-the-art across diverse scientific domains, achieving superior solution quality while significantly improving token efficiency. Our work highlights the critical role of momentum-based history in overcoming the context bottleneck for automated scientific discovery.

%% file: sections/appendix.tex
\newpage
\appendix
\onecolumn

\begin{center}
\Large\textbf{Appendix}
\end{center}

\section{Optimization vs. Algorithmic Discovery}

\paragraph{Concrete Example of Optimization and Algorithmic Discovery.}
\Cref{tab:concrete_forms} highlights the structural mismatch between classical
optimization and algorithmic discovery. While classical methods optimize
continuous parameters using analytical gradients, algorithmic discovery searches
over discrete program spaces using evaluator-based feedback. Improvements therefore
arise from semantic changes in program logic rather than numerical updates,
motivating evolution signals that explicitly capture how programs change and why
they improve.

\begin{table}[!h]
\centering
\caption{Concrete form comparison: Classical Optimization vs. Algorithmic Discovery.}
\label{tab:concrete_forms}
\resizebox{\linewidth}{!}{
\begin{tabular}{@{}l p{0.40\linewidth} p{0.48\linewidth}@{}}
\toprule
& Classical Optimization & Algorithmic Discovery \\
& \textit{(e.g., Least Squares)} & \textit{(e.g., Circle Packing)} \\
\midrule
(i) Objective & 
$\displaystyle \min_{x} \|Ax - b\|_2^2$ & \texttt{\small max sum(radius of circles)} \newline
\texttt{\small s.t. no overlap}  \\
\cmidrule{1-3}
(ii) Space & 
$x = [x_1, \dots, x_n]^\top$ & 
\texttt{\small def pack(box, n): ... return} \\
\cmidrule{1-3}
(iii) Feedback & 
$\nabla f = 2A^\top(Ax - b)$ \newline \textit{(Gradient Vector)} & 
\texttt{\small Score: 0.65} \newline \textit{(Evaluator Information)} \\
\cmidrule{1-3}
(iv) Solution & 
$x^* = [0.01, -1.2]^\top$ & 
\texttt{\small if box.w < 10: spiral()} \newline \texttt{\small else: greedy\_place()} \\
\bottomrule
\end{tabular}
}
\end{table}

\section{Algorithm of Progressive Disclosure Sampler}

\Cref{alg:multilevel_db} presents the progressive disclosure sampler that adaptively renders historical nodes at different abstraction levels to retain key evolution signals while minimizing context length.

\begin{algorithm}[h!]
\caption{Progressive Disclosure Sampler}
\label{alg:multilevel_db}
\begin{algorithmic}[1] 
    \REQUIRE Database $\mathcal{D}=\{N_1,\dots,N_T\}$, where $N_t=(\delta_{t}^{(1)},\delta_{t}^{(2)},p_t)$; \\
             Parent selection policy $\pi_{\text{select}}$; \\
             Maximum elites $k$, maximum inspirations $m$; \\
             Recent window size $w$
    \ENSURE Final prompt context $C_{\text{final}}$
    
    \STATE \hrulefill
    
    \STATE \textit{// Phase 1: Node Selection} 
    \STATE $N_{\text{parent}} \gets \mathrm{Sample}(\mathcal{D}, \pi_{\text{select}})$
    \STATE $\mathcal{N}_{\text{elite}} \gets \mathrm{TopK}(\mathcal{D}, R(p), k)$
    \STATE $\mathcal{N}_{\text{div}} \gets \mathrm{MAPElitesSample}(\mathcal{D}, \text{exclude}=N_{\text{parent}}, m)$

    \STATE 
    
    \STATE \textit{// Phase 2: Progressive Rendering}
    \STATE $C_{\text{hist}} \gets \emptyset$
    \FOR{$N_i \in (\mathcal{N}_{\text{elite}} \cup \mathcal{N}_{\text{div}})$}
        \IF{$\mathrm{Time}(N_i) > t - w$}
            \STATE $C_{\text{hist}} \gets C_{\text{hist}} \oplus \mathrm{Format}(\delta_{i}^{(2)})$ \hfill \textit{// Level 2: delta plan}
        \ELSE
            \STATE $C_{\text{hist}} \gets C_{\text{hist}} \oplus \mathrm{Format}(\delta_{i}^{(1)})$ \hfill \textit{// Level 1: summary}
        \ENDIF
    \ENDFOR
    \STATE $C_{\text{final}} \gets C_{\text{hist}} \oplus \mathrm{Format}(p_{\text{parent}})$ \hfill \textit{// Level 3: full code}

    \STATE

    \STATE \textbf{return} $C_{\text{final}}$
\end{algorithmic}
\end{algorithm}

\section{Prompt Templates}
\label{sec:prompt_template}

This appendix section provides the exact system and user prompts used by the \deltaevolve pipeline, highlighting the key architectural shifts from standard baselines like AlphaEvolve. In the \textbf{System Prompt}, we introduce a novel ``Delta Logging Instructions'' component; this explicitly constraints the LLM to not only generate code but also synthesize a structured \texttt{DELTA-SUMMARY} (Level 1) and \texttt{DELTA-PLAN} (Level 2) post-modification.

\begin{figure}[h!]
\centering
\begin{systempromptbox}

\textbf{---------------------------------  Problem Description  ---------------------------------}

\{problem\_details\}\\

\textbf{---------------------- Delta Logging Instructions \textcolor{red}{(New in DeltaEvolve)} --------------- }

You MUST summarize your changes at the very end of your response using the strict format
below.
This log is critical for the evolution memory system. Failure to follow these rules will
break the experiment.
\\

\textit{\#\#\#CRITICAL RULES for DELTA SUMMARY:}

1. NO META-TALK: Do NOT say Replace 22 lines with 230 lines, Updated code, or Changed
logic.

2. ALGORITHMIC ONLY: Describe the strategy change (e.g., Switched from Greedy to Simulated
Annealing).

3. NO TEMPLATES: Do NOT output placeholder text like high-level plan summary in one
sentence. Write actual content.

4. NO CODE SYNTAX: Do not write Python code in the summary (e.g., no def function()).
Use natural language.
\\

\textbf{\#DELTA-SUMMARY-START}

FROM: <\textit{One-sentence summary of the OLD strategy (the parent node's approach)}>

TO: <\textit{One-sentence summary of the NEW strategy (your current approach)}>

\textbf{\#DELTA-SUMMARY-END}

----------------------------------------------------------------------------------------------------------------

\textit{\#\#\#CRITICAL RULES for DELTA PLAN DETAILS}

You must explain HOW and WHY the logic changed using the strict format below.

1. Target Audience: A researcher trying to reproduce your experiment without seeing the code.

2. BE QUANTITATIVE: Do not say increased parameters. Say increased grid\_resolution from 10 to 50.

3. NAME THE ALGORITHM: Use standard terminology (e.g., Coordinate Descent, Simulated Annealing, Penalty Method).

4. NO META-TALK: Do not say I defined a function. Describe the LOGIC flow.
\\

\textbf{\#DELTA-PLAN-DETAILS-START}

[Modification 1] \\
COMPONENT:   <\textit{The specific module changed. e.g., Initialization Strategy, Constraint Handling, Optimization Loop}> \\
OLD\_LOGIC:  <\textit{Brief summary of what was removed. e.g., Random noise injection}> \\
NEW\_LOGIC:  <\textit{DETAILED mechanism. MUST include specific hyperparameters, formulas used, or heuristic rules.}> \\
HYPOTHESIS: <\textit{The scientific reasoning.}>\\

[Modification 2] (If applicable) \\
COMPONENT:   ... \\
OLD\_LOGIC:  ... \\
NEW\_LOGIC:  ... \\
HYPOTHESIS:  ...

\textbf{\#DELTA-PLAN-DETAILS-END}
\end{systempromptbox}

\caption{\textbf{System Prompt Template.} Includes novel instructions requiring the model to output structured Delta Summaries (Level 1) and Delta Plans (Level 2) alongside code, enabling automated evolutionary logging.}
\label{fig:system_prompt}
\end{figure}

The \textbf{User Prompt} is restructured to replace the token-heavy full source code of inspiration programs with a ``Delta of History Nodes'' section. Instead of raw implementation details, this section feeds the agent a concise trajectory of past successful strategies and rationales, maximizing context efficiency while providing a clear directional signal for future mutations.

\begin{figure}[h!]
\centering
\begin{userpromptbox}
\textbf{----------------------------  Current Program Information  ----------------------------}

- Focus areas: \{improvement\_areas\}

- Feedback: \{evaluator\_feedback\}

\textbf{------------ Inspirations: Delta of History Nodes \textcolor{red}{(New in DeltaEvolve)} ------------}

Below we provide history delta plans from prior nodes, including top-performing programs and highly diverse alternatives to the parent. Each delta highlights concrete strategy differences that may inspire new solution directions rather than direct reuse.
\\

\#\#\# Top Delta Plan 1

\{Delta summaries or Delta plans.\}

- Feedback: \{evaluator\_feedback\} \\

\#\#\# Top Delta Plan 2

\{Delta summaries or Delta plans.\}

- Feedback: \{evaluator\_feedback\}\\

\#\#\# Top Delta Plan 3

\{Delta summaries or Delta plans.\}

- Feedback: \{evaluator\_feedback\}\\

\#\#\# Diverse Delta Plan 1

\{Delta summaries or Delta plans.\}

- Feedback: \{evaluator\_feedback\}\\

\#\#\# Diverse Delta Plan 2

\{Delta summaries or Delta plans.\}

- Feedback: \{evaluator\_feedback\}\\

\textbf{----------------------------------- Parent Program  -----------------------------------}

\{parent\_program\}\\

Suggest improvements to the program that will improve its FITNESS SCORE.
The system maintains diversity across these dimensions: complexity, diversity
Different solutions with similar fitness but different features are valuable.

You MUST use the exact SEARCH/REPLACE diff format shown below to indicate changes:

<<<<<<< SEARCH\\
<\textit{Original code to find and replace (must match exactly)}>\\
=======\\
<\textit{New replacement code}>\\
>>>>>>> REPLACE\\

\end{userpromptbox}
\caption{\textbf{User Prompt Template.} Replaces full-code inspiration contexts with concise Delta Summaries and Plans, exposing evolutionary trajectories while minimizing token usage.}
\label{fig:user_prompt}
\end{figure}

\newpage
\section{Task Details}
\label{sec:task_details}

\subsection{Blackbox Optimization}

\paragraph{Task Description} The Blackbox Optimization task involves minimizing continuous objective functions drawn from the BBOB (Black-Box Optimization Benchmarking) suite, which encompasses a diverse range of landscapes including separable, ill-conditioned, and multi-modal functions. The optimizer must operate as a true black box, receiving only the problem dimension, bounds, and evaluation budget without access to gradients or the analytical form of the function. The objective is to locate the global optimum with high precision and efficiency across various function identifiers and instance realizations, while strictly adhering to search space bounds and ensuring reproducibility via fixed random seeds.

\paragraph{Initial Programs} The initial program implements a baseline Random Search algorithm that explores the solution space by sampling candidate points uniformly within the hyperbox defined by the problem bounds(range of 
$[-5,5]$). It dynamically detects the problem dimensionality and employs a simple "keep-best" logic, where the solution with the lowest observed function value is retained. This approach is purely stochastic and memory-less, serving as a fundamental baseline that does not adapt any search strategy.

\paragraph{Evaluator} We evaluated the optimizer on five BBOB functions with increasing complexity: \texttt{sphere\_d3\_i1}, \texttt{rosenbrock\_d5\_i2}, \texttt{rastrigin\_d10\_i5}, \texttt{ellipsoid\_d20\_i1}, and \texttt{schaffers\_d40\_i5}. The identifiers follow the format \texttt{<function name>\_<input dimension>\_<instance ID>}. Reference values ($V_{\text{ref}}$) were computed using \texttt{scipy.optimize}. The evaluation employs a two-stage protocol: Stage 1 acts as a validity filter, allowing only successful runs to proceed to Stage 2 for final scoring. The final score combines solution quality and efficiency:
\begin{equation}
    S_{\text{case}} = 0.7 \cdot S_{\text{val}} + 0.3 \cdot \max\left(0, 1 - \frac{N_{\text{used}}}{N_{\text{budget}}}\right)
\end{equation}
The value score $S_{\text{val}}$ depends on the normalized improvement $\delta = (V_{\text{ref}} - V_{\text{best}})/|V_{\text{ref}}|$, calculated as $1+\delta$ if $V_{\text{best}} \le V_{\text{ref}}$ (better or equal), and $(1+|\delta|)^{-1}$ otherwise.

\begin{figure}[h]
    \centering
    \begin{minipage}{0.49\textwidth}
        \centering
        \includegraphics[width=\linewidth]{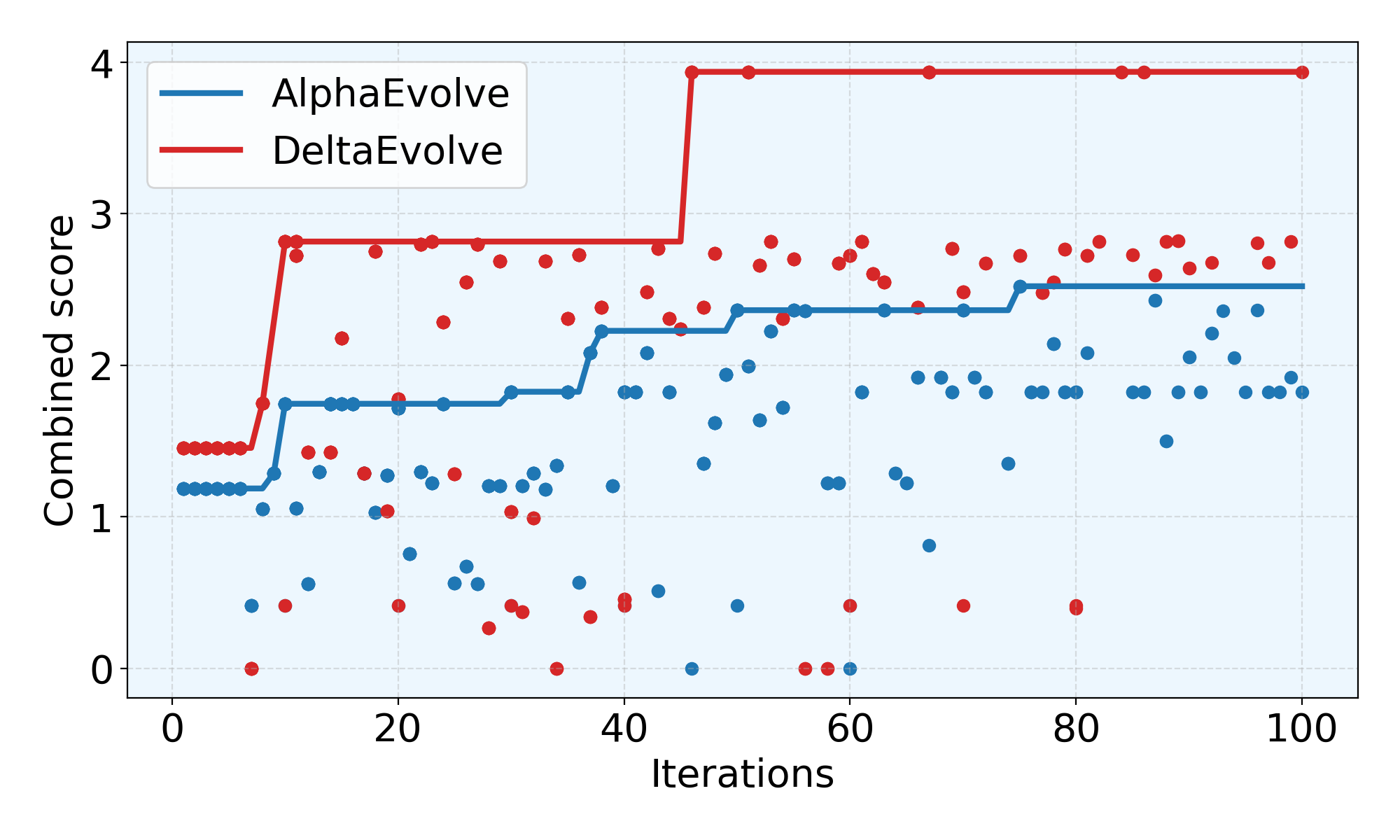}
    \end{minipage}
    \hfill
    \begin{minipage}{0.49\textwidth}
        \centering
        \includegraphics[width=\linewidth]{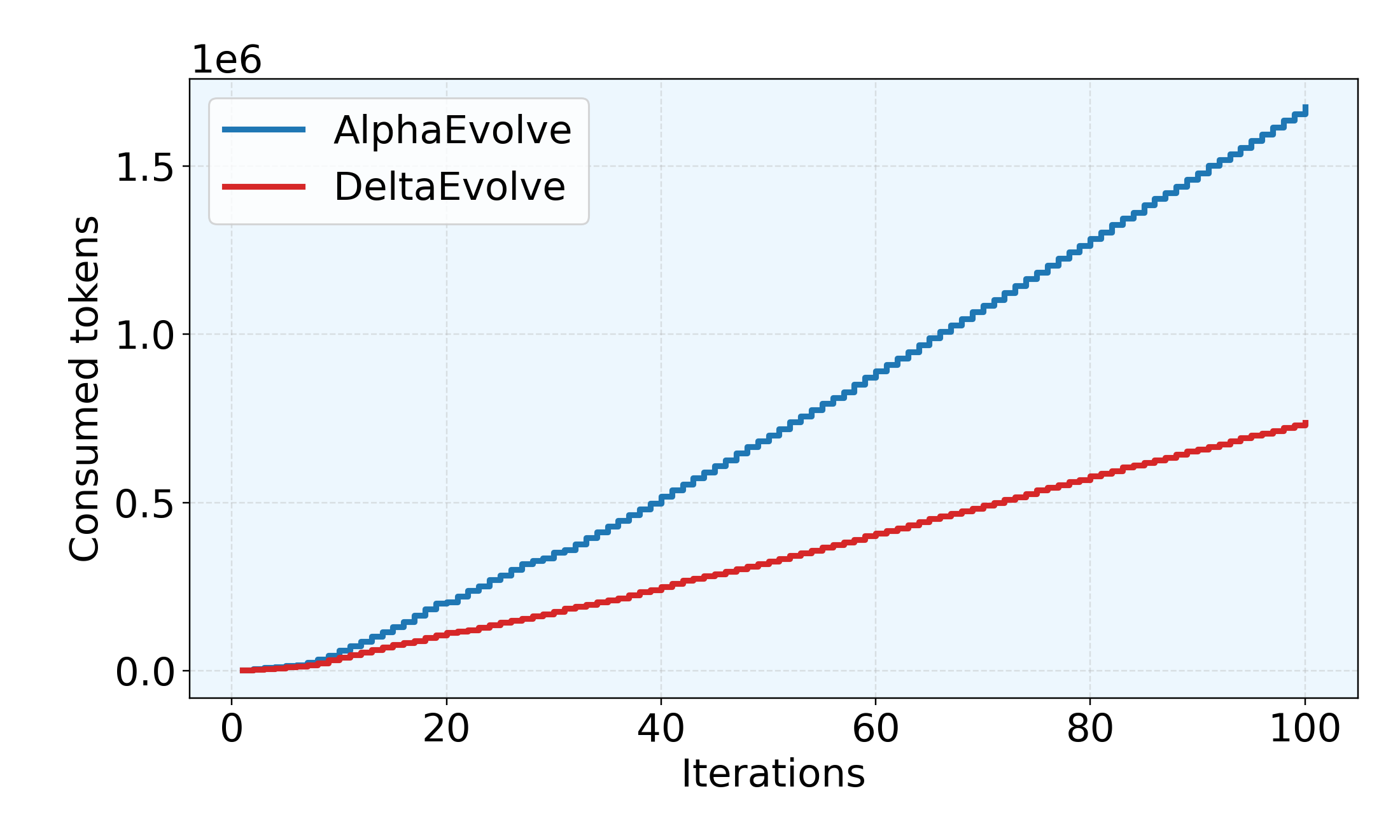}
    \end{minipage}
    \caption{Evolutionary trajectory for a representative run. The left panel shows the score progression, and the right panel tracks token usage per iteration. \deltaevolve reaches higher performance with reduced token cost.}
    \label{fig:bbob_evolution}
\end{figure}

\paragraph{Evolution Process} To clearly illustrate the evolutionary dynamics, we select a representative run using a fixed random seed ($42$). Figure~\ref{fig:bbob_evolution} visualizes the optimization process, comparing the score of the combined score (left) and the cumulative token consumption (right) across iterations. The results demonstrate that \deltaevolve achieves a significantly higher final score while maintaining lower token usage compared to AlphaEvolve, highlighting its superior efficiency in exploring the code space.

\subsection{Hexagon Packing}

\paragraph{Task Description} The Hexagon Packing task requires constructing an optimal arrangement of $N=11$ disjoint unit regular hexagons (side length $s=1$) within a minimal bounding regular hexagon. The objective is to maximize the inverse of the outer hexagon's side length, $\rho = 1/R$, which is equivalent to maximizing the packing density. The optimizer outputs the coordinates $(x_i, y_i)$ and rotation angles $\theta_i$ for all $N$ inner hexagons, as well as the geometry of the outer hexagon. A valid solution must satisfy strict geometric constraints: all pairs of inner hexagons must be disjoint, and all inner hexagons must be fully contained within the outer boundary.

\paragraph{Initial Programs} The initial program implements a deterministic, heuristic approach. It arranges the 11 unit hexagons in a static, pre-calculated grid pattern (a ``lattice" layout) centered at the origin. The outer hexagon side length $R$ is set to a conservative value ($R=8$) to guarantee containment. While valid, this solution is sparse and far from optimal, serving as a lower-bound baseline for the packing density.

\paragraph{Evaluator} The evaluator validates the geometric feasibility of the proposed packing using the Separating Axis Theorem (SAT). It performs $O(N^2)$ pairwise checks to ensure disjointness and $O(N)$ checks for containment within the outer boundary. If any constraint is violated (overlap or protrusion) within a numerical tolerance of $10^{-6}$, the solution is marked as invalid (score 0). Valid solutions are scored based on the inverse radius $\rho = 1/R$, normalized against a state-of-the-art benchmark ($\rho_{\text{ref}} \approx 0.2544$). The final metric balances the packing density and the computational time required to generate the solution.

\begin{figure}[h]
    \centering
    \begin{minipage}{0.49\textwidth}
        \centering
        \includegraphics[width=\linewidth]{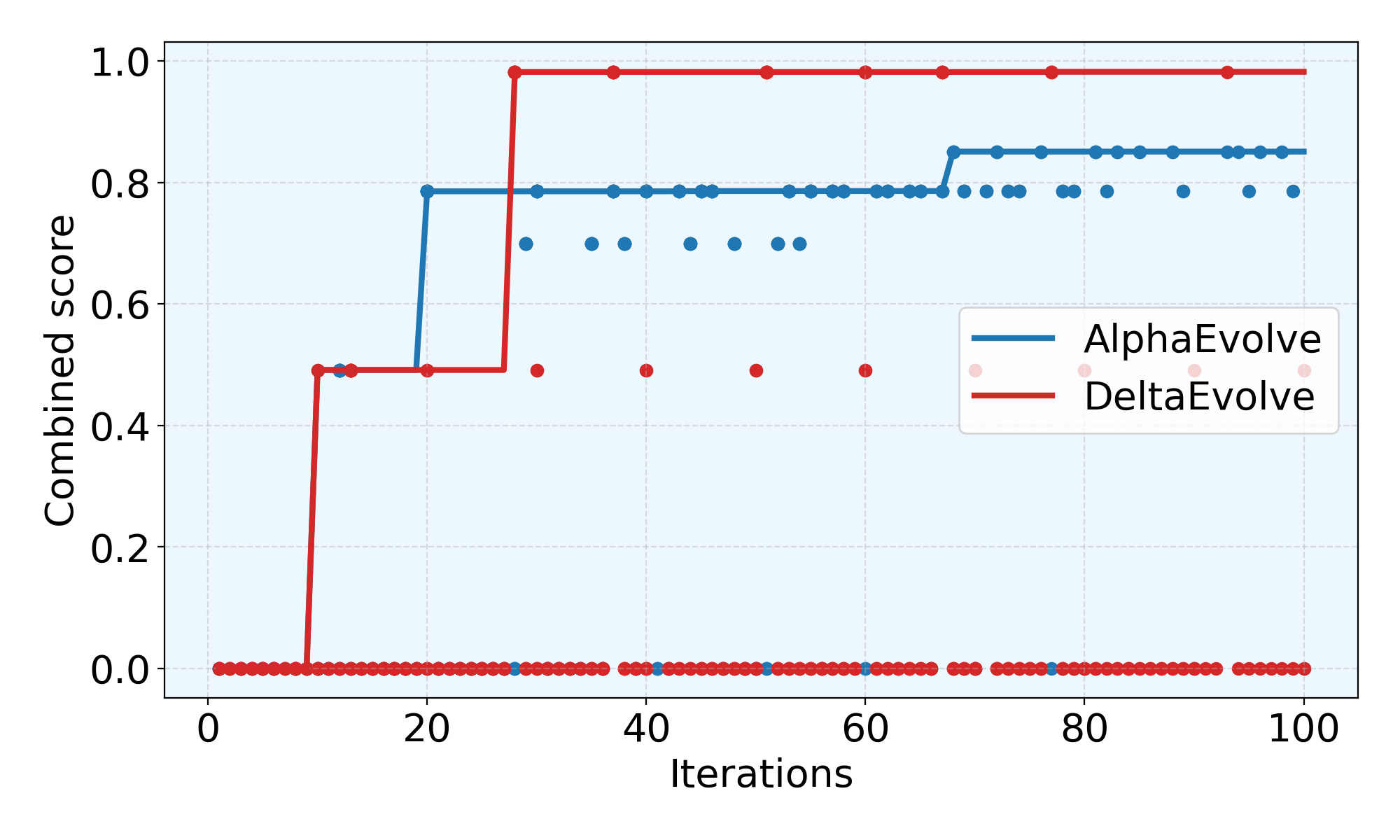}
    \end{minipage}
    \hfill
    \begin{minipage}{0.49\textwidth}
        \centering
        \includegraphics[width=\linewidth]{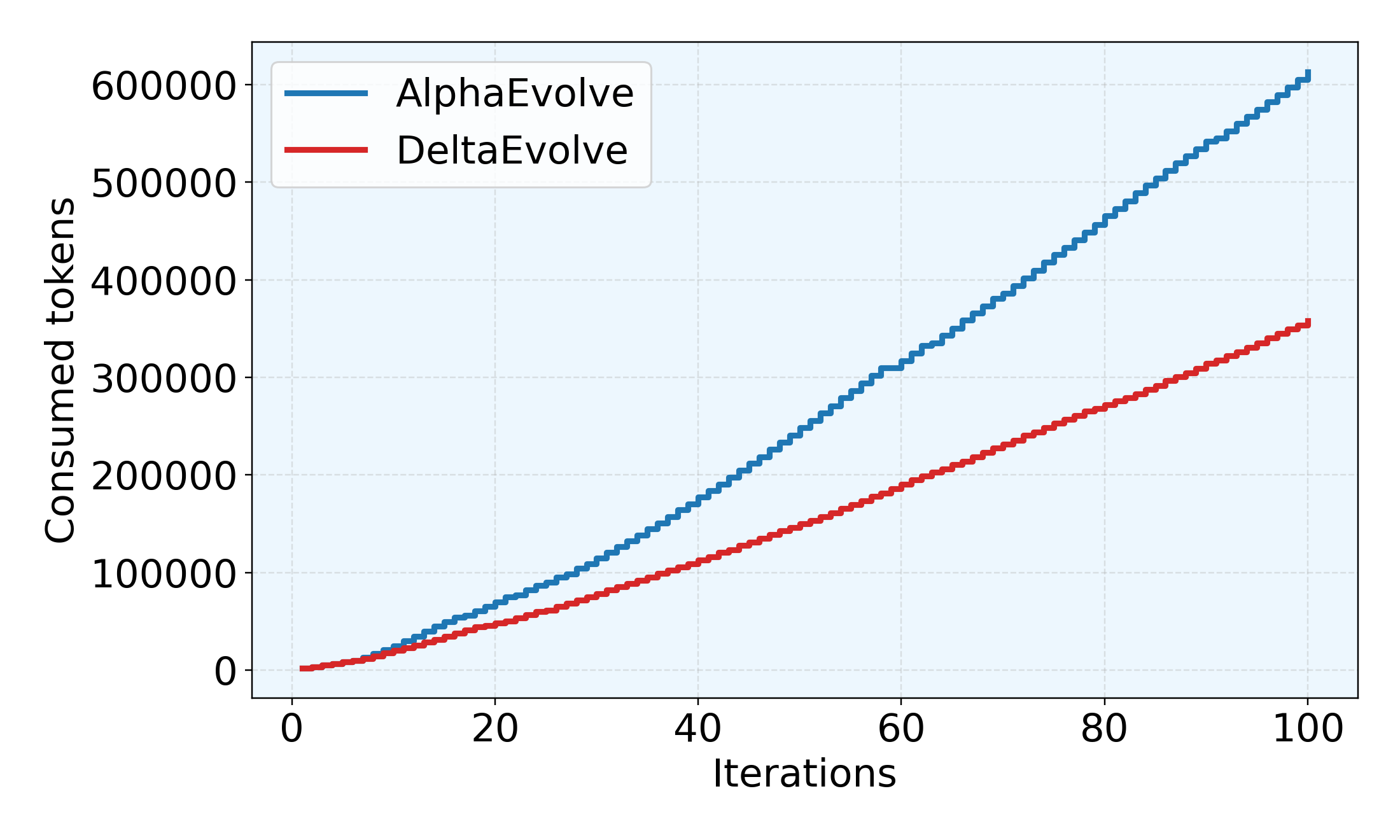}
    \end{minipage}
    \caption{Optimization trajectory for the Hexagon Packing task. The left panel illustrates the improvement, while the right panel shows the token efficiency of the search process.}
    \label{fig:hex_evolution}
\end{figure}

\paragraph{Evolution Process} To clearly illustrate the evolutionary dynamics, we select a representative run using a fixed random seed (42). Figure~\ref{fig:hex_evolution} displays the progression of the packing density score (left) and token consumption (right). \deltaevolve rapidly moves beyond the initial lattice configuration, employing a continuous relaxation strategy that iteratively resolves overlaps while compressing the outer boundary. This allows it to approach the theoretical packing limit more closely and efficiently than the baseline random search.

\subsection{Symbolic Regression}

\paragraph{Task Description}
The Symbolic Regression task involves discovering a mathematical expression that accurately models a physical process based on observational data. Specifically, the objective is to predict the acceleration ($dv/dt$) of a nonlinear harmonic oscillator given input features: position ($x$), time ($t$), and velocity ($v$). The optimizer must evolve the functional form of a Python function \texttt{func(x, params)}, where \texttt{x} is the input matrix and \texttt{params} is a vector of learnable coefficients (up to 10). This constitutes a hybrid optimization problem: the evolutionary algorithm searches for the optimal symbolic structure (e.g., polynomial terms, trigonometric functions), while a numerical optimizer adjusts the coefficients to fit the data.

\paragraph{Initial Programs}
The initial program implements a naive linear model: $f(\mathbf{x}, \mathbf{p}) = p_0 \cdot x + p_1 \cdot t + p_2 \cdot v$. While computationally efficient and easy to optimize, this model is fundamentally insufficient for capturing the dynamics of a nonlinear oscillator, which typically requires restoring forces (dependent on position) and damping terms (dependent on velocity) that interact non-linearly (e.g., cubic stiffness $x^3$ in a Duffing oscillator).

\paragraph{Evaluator}
The evaluation pipeline measures the predictive accuracy of the proposed model on a training dataset. For every candidate functional structure, the evaluator employs the BFGS algorithm (\texttt{scipy.optimize.minimize}) to optimize the vector \texttt{params} by minimizing the Mean Squared Error (MSE) between the predicted acceleration and the ground truth. The model's fitness is quantified as $-\log_{10}(\text{MSE})$, promoting high-precision solutions. The evaluator also enforces robustness checks, assigning penalizing scores to models that generate numerical errors (e.g., \texttt{NaN}, overflow) or fail to execute within a strictly defined timeout.

\begin{figure}[h]
    \centering
    \begin{minipage}{0.49\textwidth}
        \centering
        \includegraphics[width=\linewidth]{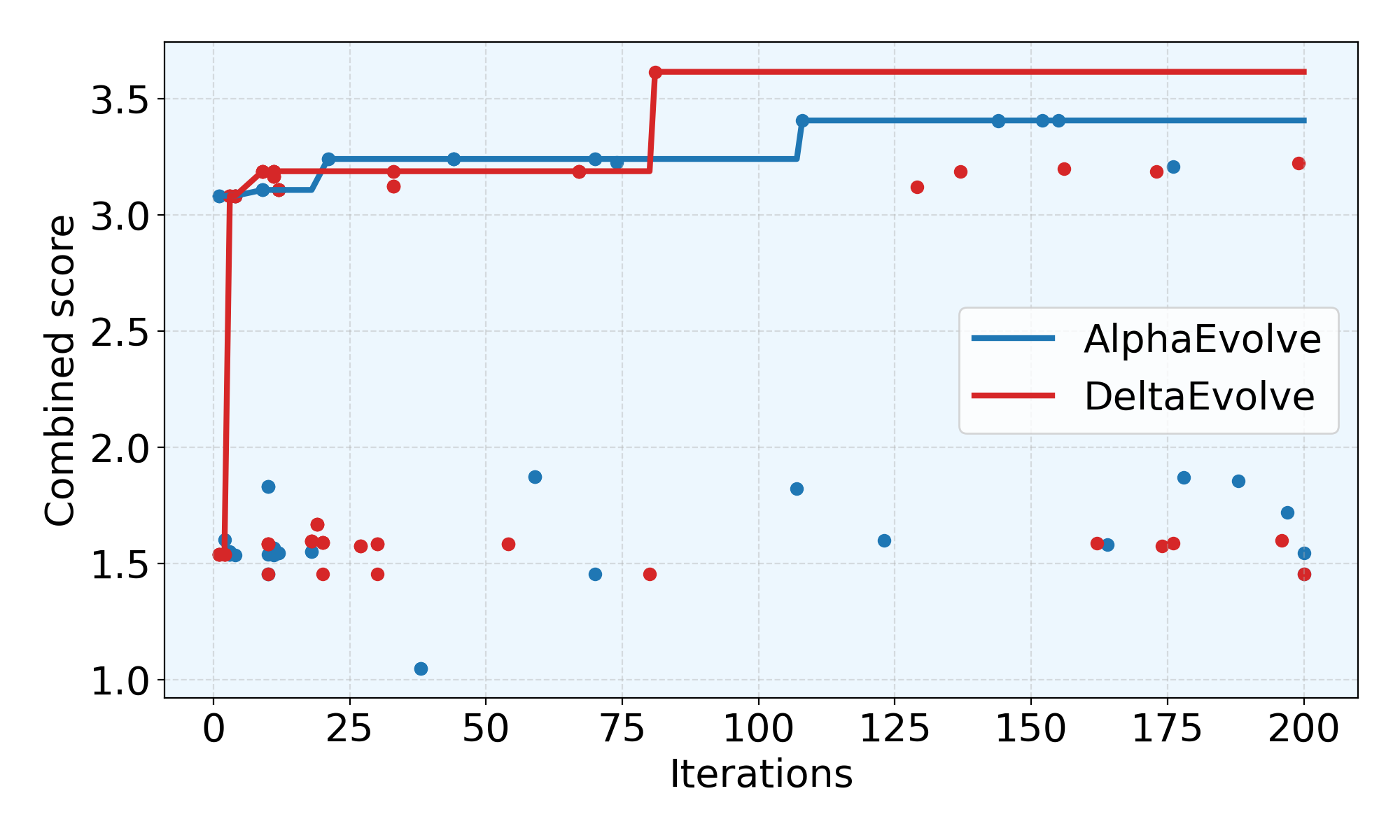}
    \end{minipage}
    \hfill
    \begin{minipage}{0.49\textwidth}
        \centering
        \includegraphics[width=\linewidth]{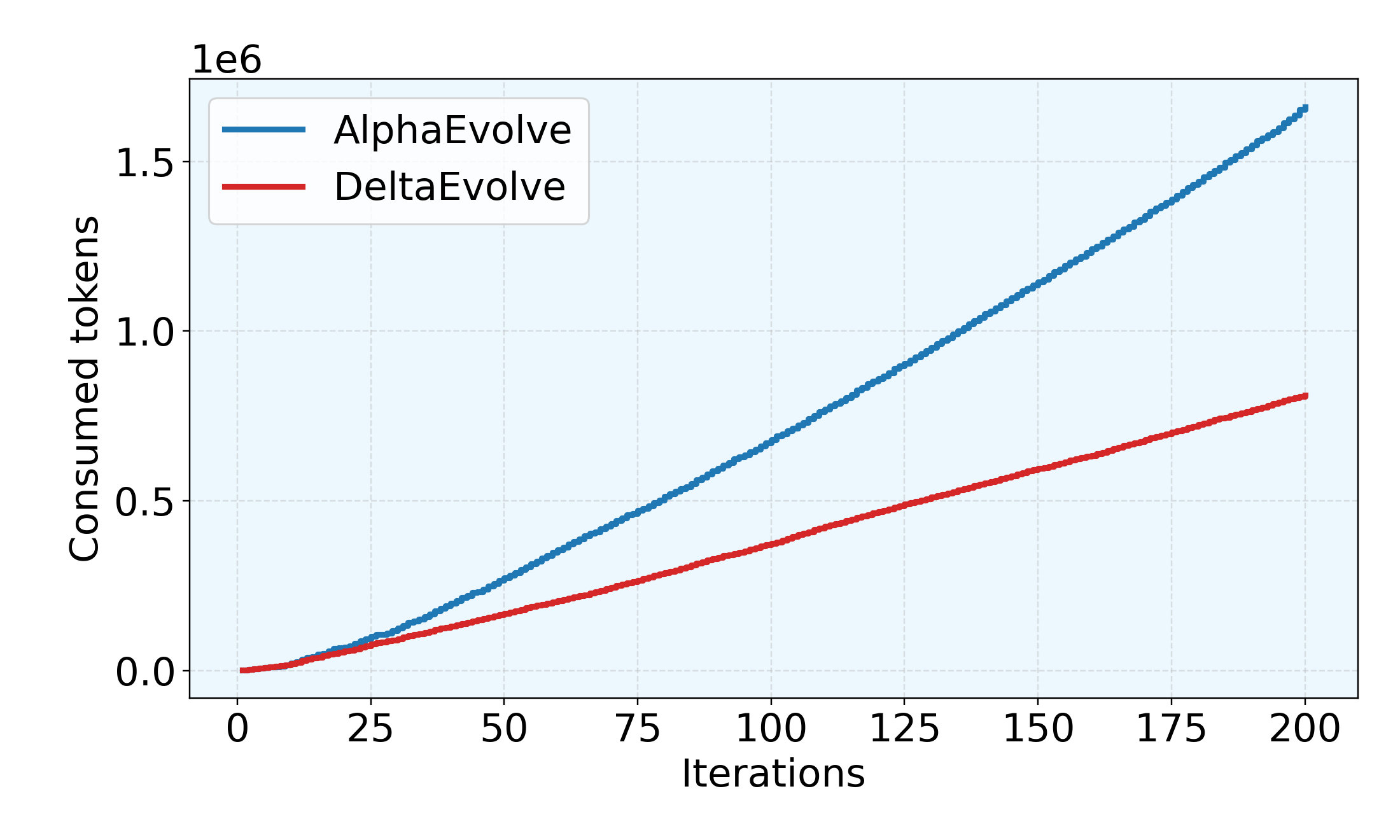}
    \end{minipage}
    \caption{Evolutionary progress for Symbolic Regression. The left panel shows the score, and the right panel tracks token usage. The optimizer quickly identifies the necessary nonlinear terms to fit the physical dynamics.}
    \label{fig:symreg_evolution}
\end{figure}

\paragraph{Evolution Process}
We analyze the search trajectory using a fixed random seed. Figure~\ref{fig:symreg_evolution} depicts the improvement in the model's score (left) and the token consumption (right) over iterations. \deltaevolve successfully transitions from the linear baseline to a non-linear formulation—identifying critical terms like cubic stiffness or interaction effects—resulting in a sharp reduction in MSE and a corresponding increase in the logarithmic score, all while maintaining concise code expressions.

\subsection{PDE Solver}

\paragraph{Task Description} The objective is to optimize a Krylov subspace solver for sparse linear systems of the form $Ax = b$. The specific focus is on the Conjugate Gradient (CG) method, which is the standard solver for Symmetric Positive Definite (SPD) systems (e.g., discretized Poisson equations). The optimizer must enhance the solver to reduce the number of iterations required for convergence and improve numerical stability against ill-conditioned matrices.

\paragraph{Initial Programs}
The initial program implements a standard Conjugate Gradient algorithm with a basic Jacobi (diagonal) preconditioner. It computes the inverse diagonal $M^{-1} = \text{diag}(A)^{-1}$ to scale the residual. While computationally cheap, the Jacobi preconditioner provides only modest improvements in convergence for coupled systems like the Poisson equation. The baseline implementation lacks restart mechanisms, making it susceptible to loss of orthogonality in high-dimensional or ill-conditioned problems.

\paragraph{Evaluator} The evaluator assesses the solver on the 2D Poisson equation discretized on square grids of sizes $N \in \{50, 100, 200\}$. This generates sparse SPD matrices with condition numbers scaling as $O(N^2)$. Performance is scored based on a composite metric of convergence speed (number of iterations), solution accuracy ($L_2$ error relative to ground truth), and the final residual norm. The score penalizes slow convergence exponentially: $S \propto e^{-k/N}$, encouraging solvers that drastically reduce the iteration count $k$.

\begin{figure}[h]
    \centering
    \begin{minipage}{0.49\textwidth}
        \centering
        \includegraphics[width=\linewidth]{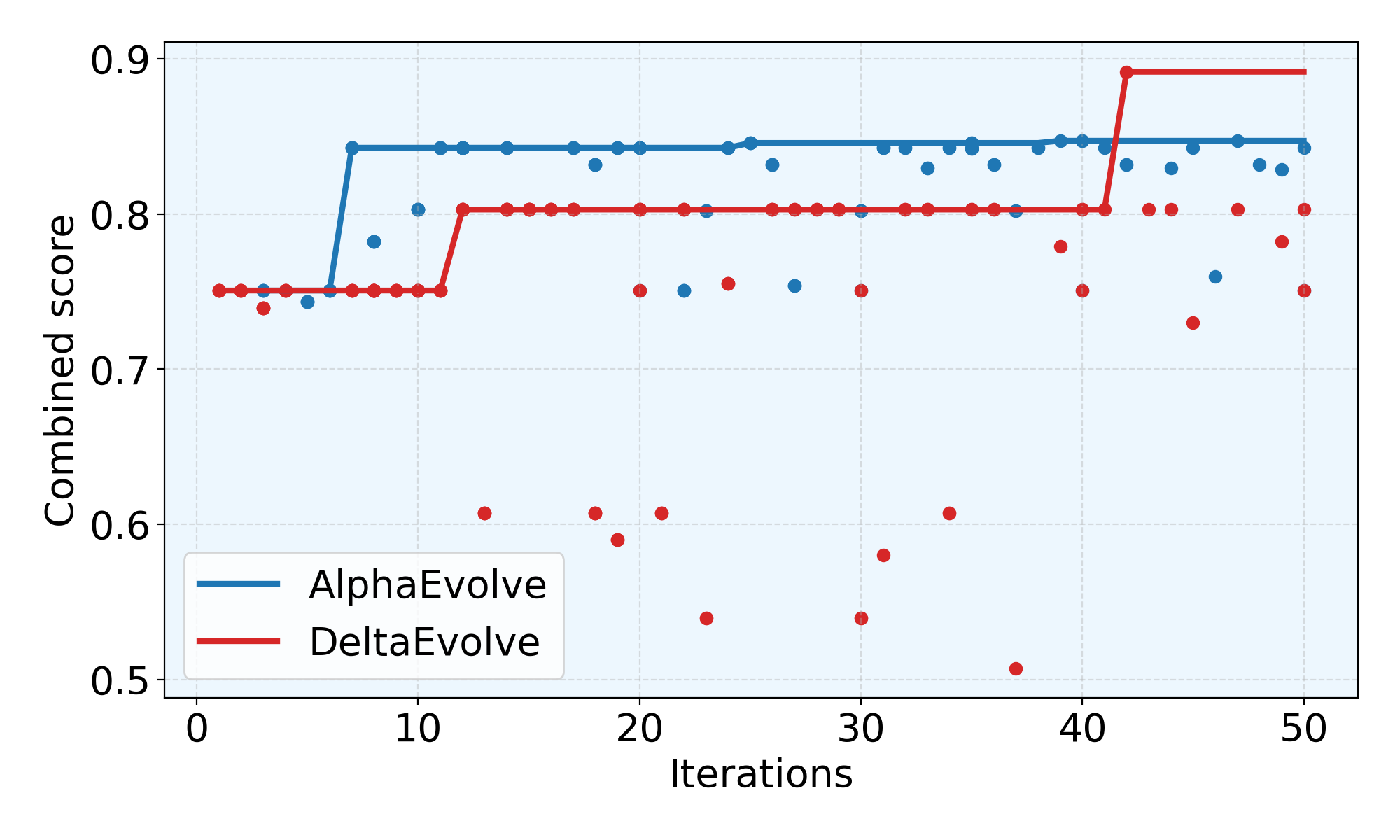}
    \end{minipage}
    \hfill
    \begin{minipage}{0.49\textwidth}
        \centering
        \includegraphics[width=\linewidth]{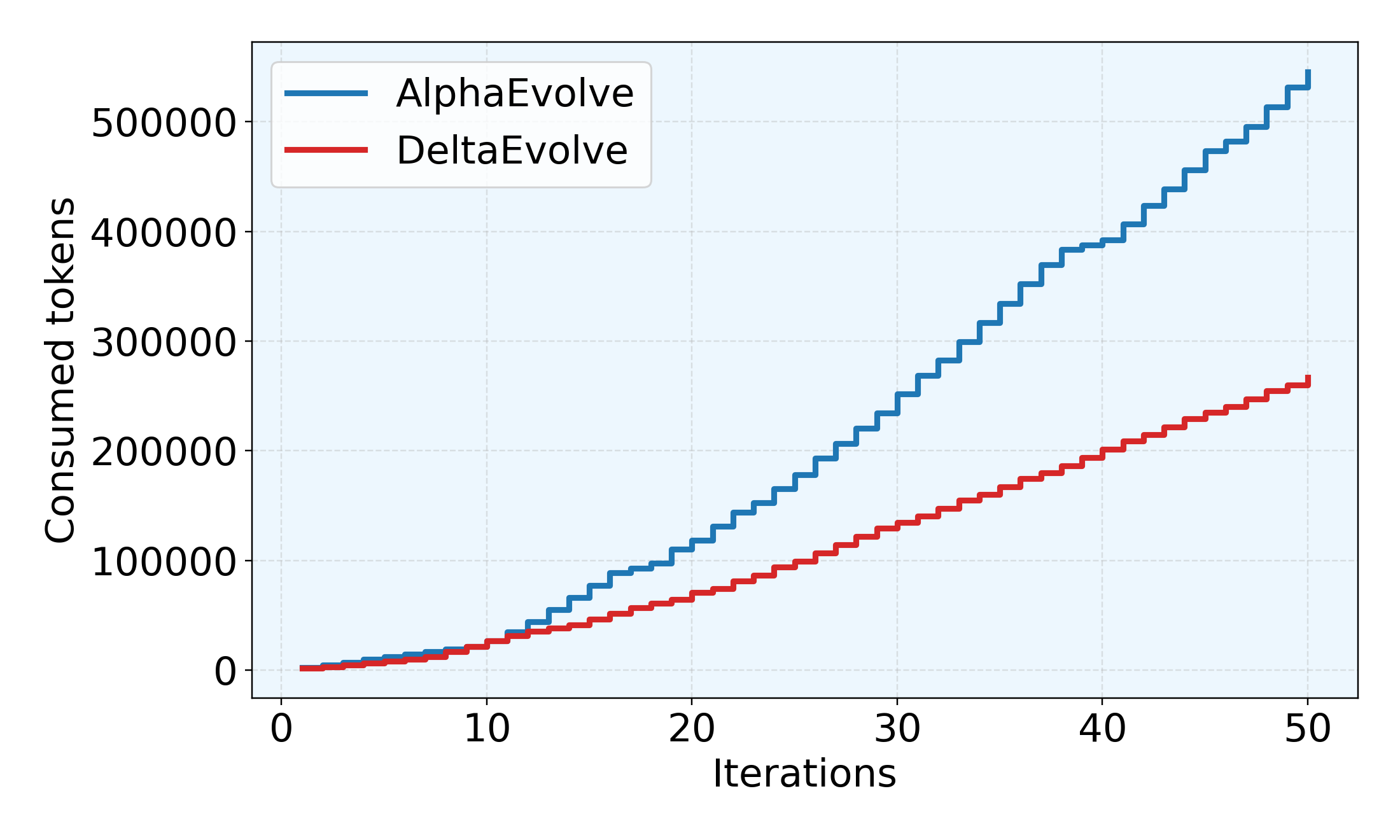}
    \end{minipage}
    \caption{Optimization trajectory for the PDE Solver. The left panel tracks the combined accuracy/speed score, and the right panel tracks token usage. The \deltaevolve achieves better combined score and less token usage.}
    \label{fig:pde_evolution}
\end{figure}

\paragraph{Evolution Process} We visualize the evolution using a fixed random seed. Figure~\ref{fig:pde_evolution} compares the combined score (left) and token usage (right). \deltaevolve rapidly identifies that the baseline Jacobi preconditioner is the bottleneck. It evolves a more sophisticated Polynomial Preconditioner (approximating $A^{-1}$ via a Neumann series) and introduces a residual-based restart heuristic. This results in a solver that converges in significantly fewer iterations than the baseline, achieving a higher score with efficient code usage.

\subsection{Efficient Convolution.}

\paragraph{Task Description} This task focuses strictly on computational efficiency. It involves optimizing a 2D convolution operation, a fundamental kernel in signal processing and computer vision. The inputs are two real-valued matrices: a large input array $A$ of size $(30n \times 30n)$ and a smaller kernel $B$ of size $(8n \times 8n)$, where $n$ is a scaling factor. The operation is defined in ``full'' mode with zero-padding (``fill'' boundary handling), meaning the output size is $(H_A + H_B - 1) \times (W_A + W_B - 1)$. The objective is to maximize the speedup relative to a reference \texttt{scipy.signal.convolve2d} implementation while maintaining numerical correctness ($L_2$ relative error $< 10^{-6}$).

\paragraph{Initial Programs}
The initial program is a direct wrapper around \texttt{scipy.signal.convolve2d}, providing a functional baseline but lacking any specific optimizations for the problem structure. It operates in the spatial domain, which scales as $O(N_A^2 N_B^2)$, becoming computationally prohibitive for large $n$. The baseline does not leverage parallelization, vectorization, or frequency-domain transformations that could significantly accelerate the computation.

\paragraph{Evaluator}
The evaluator benchmarks the evolved solution against the reference \texttt{AlgoTune} baseline. The performance metric is the \texttt{speedup ratio}: $\text{Speedup} = T_{\text{baseline}} / T_{\text{evolved}}$. Correctness is verified by comparing the Frobenius norm of the difference between the evolved output and the reference output. The evaluation pipeline includes warmup runs to stabilize JIT compilers and multiple timing trials to ensure robust measurements. Solutions that exceed a timeout or fail the accuracy check receive a zero score.

\begin{figure}[h]
    \centering
    \begin{minipage}{0.49\textwidth}
        \centering
        \includegraphics[width=\linewidth]{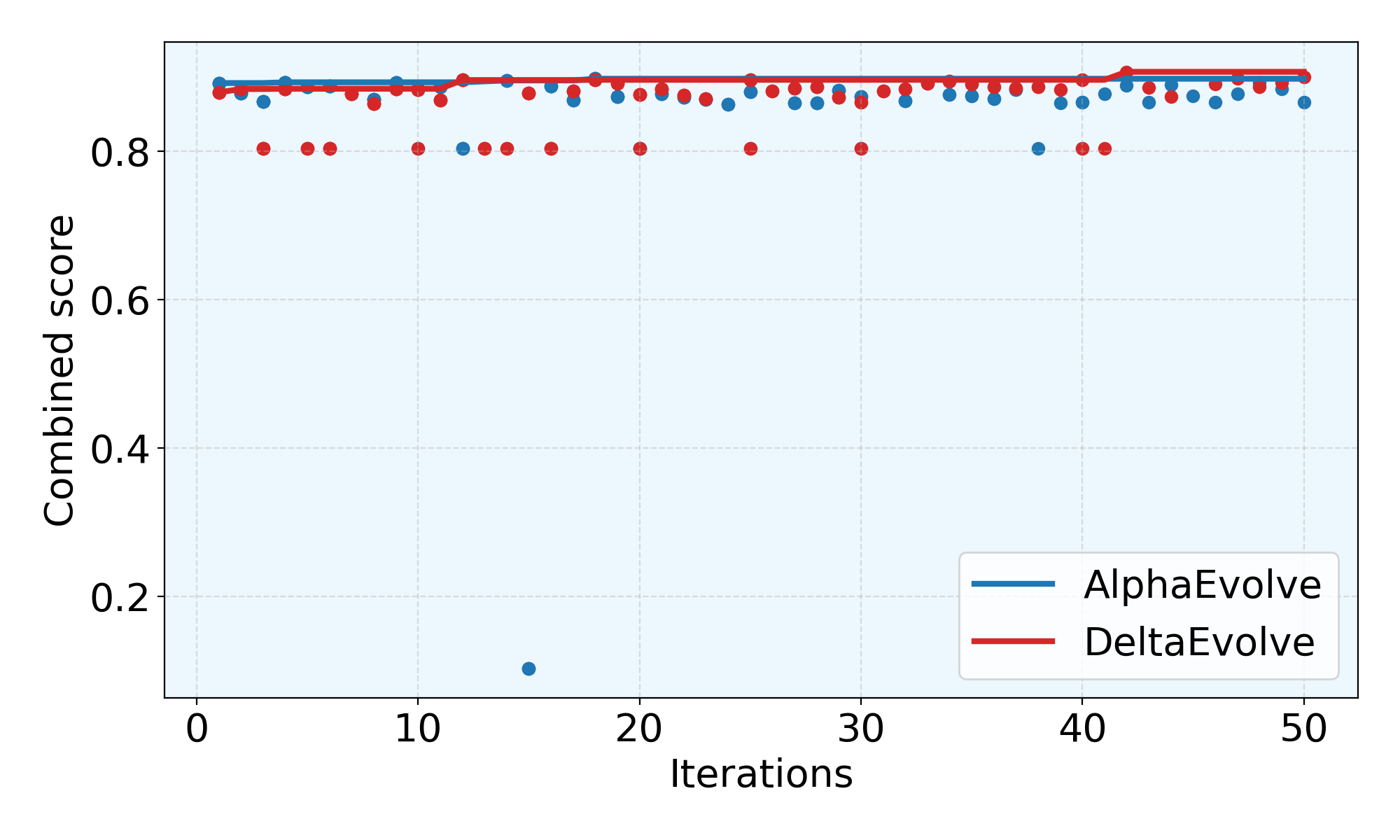}
    \end{minipage}
    \hfill
    \begin{minipage}{0.49\textwidth}
        \centering
        \includegraphics[width=\linewidth]{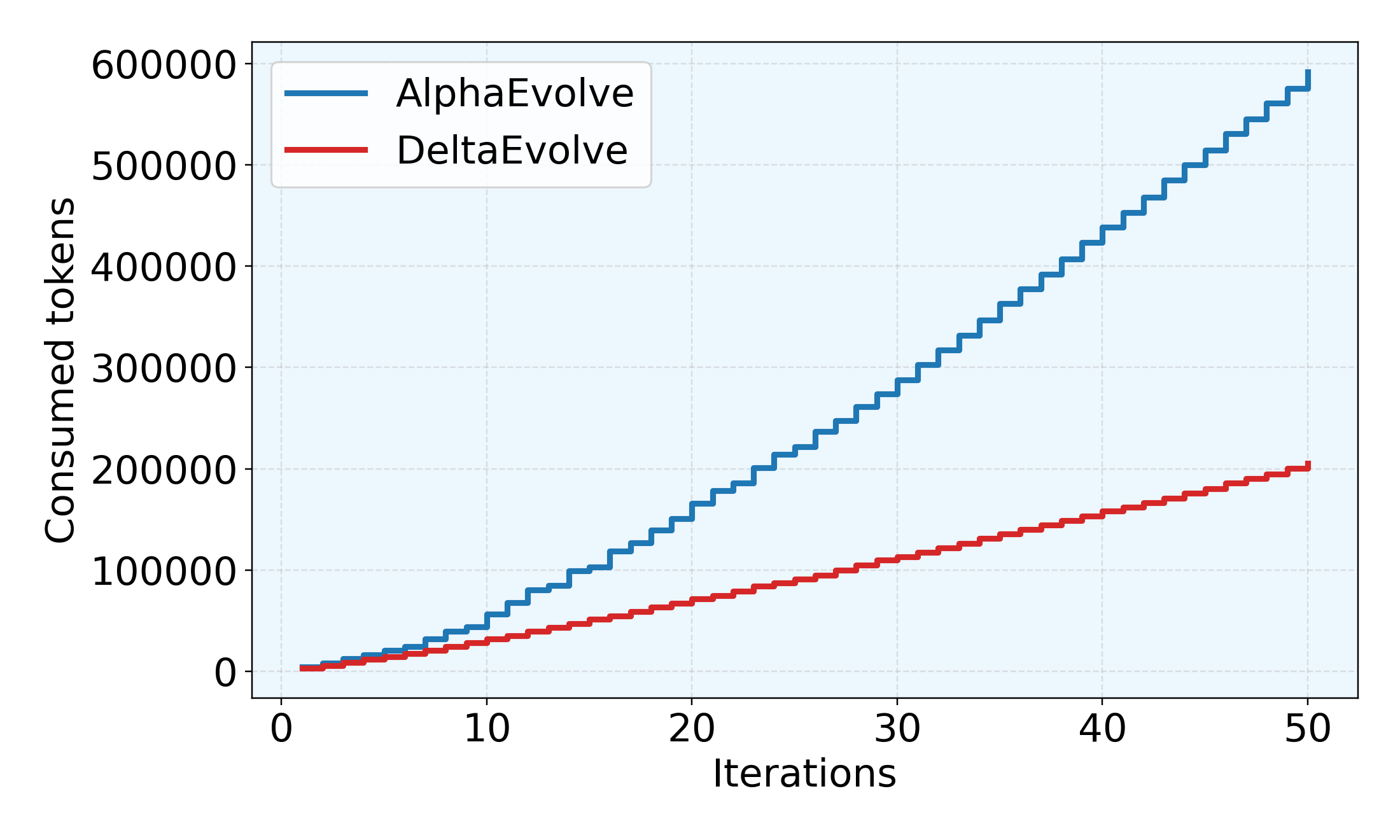}
    \end{minipage}
    \caption{Optimization trajectory for the Convolve2D task. The left panel tracks the efficiency score (speedup), while the right panel tracks token usage. \deltaevolve identifies the optimal FFT approach, significantly outperforming the baseline.}
    \label{fig:conv_evolution}
\end{figure}

\paragraph{Evolution Process}
We analyze the optimization trajectory using a fixed random seed. Figure~\ref{fig:conv_evolution} illustrates the speedup achieved (left) and the token consumption (right). \deltaevolve successfully transitions from the spatial domain baseline to a frequency-domain approach using the Fast Fourier Transform (FFT). By exploiting the Convolution Theorem, it reduces the complexity to $O(N^2 \log N)$, yielding a dramatic speedup for large inputs. It further optimizes memory layout and FFT planning to maximize efficiency.

\section{Case Study}

\begin{figure}[h!]
    \centering
    \includegraphics[width=1\linewidth]{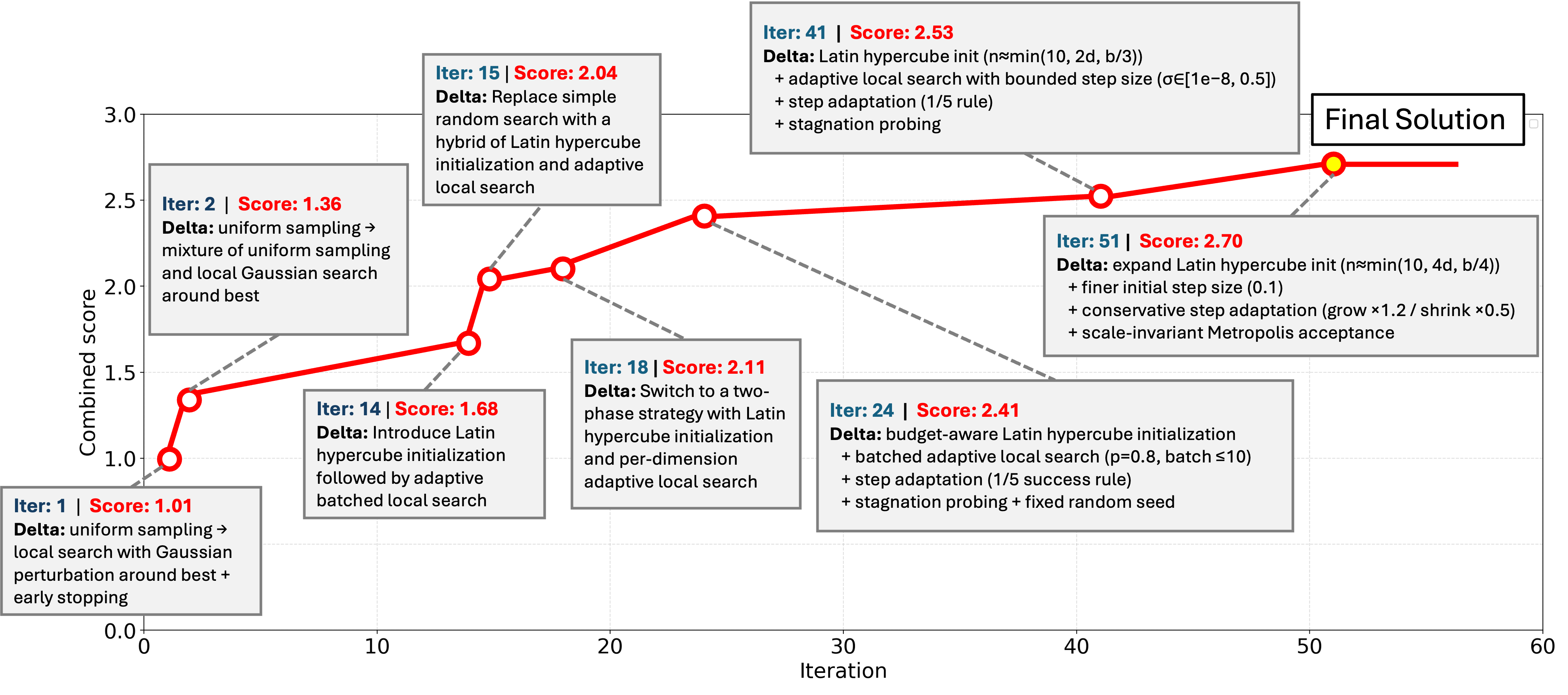}
    \caption{\textbf{Case Study of DeltaEvolve Dynamics on Black-Box Optimization.} We plot the best-so-far combined score (higher is better) versus iterations. Annotated nodes show the \textbf{Delta Summary (Level 1)} between successive solutions, illustrating how changes in initialization, adaptive local search, and acceptance mechanisms are progressively introduced.}
    \label{fig:case_study}
\end{figure}

To demonstrate \deltaevolve in practice, we present a case study on a black-box optimization task that visualizes the full evolutionary trajectory across generations. \Cref{fig:case_study} plots the best-so-far combined score (higher is better) against the number of iterations, annotating key milestones with their corresponding Level 1 Delta Summaries. The trajectory reveals a coherent evolutionary logic: initially, the agent explores basic heuristics like local Gaussian perturbations (Iter 1--2). A critical structural breakthrough occurs at Iteration 14, where the agent explicitly proposes a semantic shift to ``Latin Hypercube initialization with adaptive batched local search,'' triggering a steep performance gain ($1.68 \rightarrow 2.11$). Subsequently, rather than randomly rewriting the codebase, the agent leverages the delta history to perform targeted refinements---introducing ``stagnation probing'' (Iter 41) and ``Metropolis acceptance'' (Iter 51) to optimize budget usage and escape local optima. This progression confirms that \deltaevolve actively maintains evolutionary momentum, iteratively stacking meaningful logical improvements to converge on a robust final solution.

\newpage
\section{Examples of Semantic Deltas}
\label{sec:delta_examples}

To provide concrete intuition for our multi-level representation, we present real-world artifacts generated by \deltaevolve during the Black-Box Optimization task. \Cref{fig:level1_example} displays a \textbf{Level 1 Delta Summary}, which captures the high-level strategic pivot in a concise \texttt{FROM/TO} format—summarizing the transition from random sampling to a two-phase budget-aware optimizer without unnecessary detail. Complementing this, \Cref{fig:level2_example} shows the corresponding \textbf{Level 2 Delta Plan Details}, which acts as a structured scientific log. It breaks down the evolution into specific components (e.g., Initialization, Step-size), explicitly contrasting \texttt{OLD\_LOGIC} with \texttt{NEW\_LOGIC} and providing the underlying hypothesis. These readable logs allow the agent to perceive the causal mechanism of improvement without parsing the full source code.

\begin{figure}[h!]
\centering
\begin{level1box}
\textbf{[iter: 027 | improved]}

\textbf{FROM}: Uniform random sampling with 
occasional fixed-size local perturbations (5 Gaussian tries at 5\% range) and greedy replacement.\\

\textbf{TO}: A two-phase budget-aware optimizer: Latin Hypercube initialization (n\_init =
min(max(10,4*dim), max(1,budget//3))) followed by an adaptive batched per-dimension Gaussian
local search (sigma init = 0.2*span, grow * 1.2 on success, 
shrink *0.5 on stagnation with patience = max(3,dim), batch = min(10, remaining//10), 
exploitation p=0.8) plus occasional global probes and strict budget accounting.\\

\textbf{Evaluation Results}:\{ stage1\_cases: \{   "sphere\_d3\_i1": 0.699 \} | stage2\_cases: \{   "rosenbrock\_d5\_i2": 0.703,   "rastrigin\_d10\_i5": 4.354,   "ellipsoid\_d20\_i1": 1.270,   "schaffers\_d40\_i5": 0.700 \} \}
\end{level1box}
\caption{\textbf{Example of Delta Summary (Level 1).} Generated during Black-Box Optimization, this summary captures the high-level strategic pivot (FROM random sampling TO adaptive LHS) in a concise, token-efficient format suitable for long-term history retrieval.}
\label{fig:level1_example}
\end{figure}

\begin{figure}[h!]
\centering
\begin{level2box}
\textbf{\#\# Program Top 1:}

\textbf{[iter:038:improved]}

[Delta plan details]:\\

\textbf{[Modification 1]}

\textbf{COMPONENT}:   Initialization Strategy (Latin Hypercube)

\textbf{OLD\_LOGIC}:   n\_init = min(max(10, 4*dimension), max(1, budget//4)); per-dimension shuffle of midpoints with jitter = (rng.random - 0.5)/n\_init.

\textbf{NEW\_LOGIC}:   n\_init = min(max(10, 4*dimension), max(1, budget//3)), then clamp n\_init < max(1, budget-2) to leave two evaluations. Construct full LHS by computing base midpoints base[i] = (i+0.5)/n\_init, generating a random permutation idx per-dimension, forming perms[:,d] = base[idx], then adding jitter = (rng.random((n\_init,dimension))-0.5)/(n\_init*3.0) and clipping to [0,1]. Map to box via lower + sample*span.

\textbf{HYPOTHESIS}:  Full-permutation LHS with slightly more initialization samples improves space-filling coverage; smaller jitter reduces boundary bias while keeping diversity; leaving two evals ensures immediate subsequent polish/search.\\

\textbf{[Modification 2]}

\textbf{COMPONENT}:   Initialization Polishing (Coordinate Refinement)

\textbf{OLD\_LOGIC}:   Single small-radius sweep: refine\_evals = min(budget-evals, min(2*dim,10)), radius = 0.05*span, try +/- on each coordinate once.

\textbf{NEW\_LOGIC}:   Multi-scale polish with refine\_evals = min(budget-evals, min(3*dim,15)). Iterate scales = [0.2, 0.05, 0.01]; for each scale, compute radius = scale*span and for each dimension try +/- radius while budget remains; accept and update best, and bias current\_x/current\_val to improvements immediately.

\textbf{HYPOTHESIS}:  Many BBOB functions are separable or have strong one-dimensional curvature; coarse-to-fine coordinate probes (0.2 then 0.05 then 0.01) cheaply discover large improvements early and avoid wasting many evaluations at a single tiny scale.\\

\textbf{[Modification 3]}

\textbf{COMPONENT}:   Step-size Initialization and Lower Bound

\textbf{OLD\_LOGIC}:   initial\_step = 0.1*span; min\_step = 1e-8*span; grow=1.3/shrink=0.7.

\textbf{NEW\_LOGIC}:   initial\_step = 0.2*span; min\_step = 1e-6*span (keep grow=1.3, shrink=0.7); clamp step updates as before but with the new numeric bounds.

\textbf{HYPOTHESIS}:  A slightly larger initial step accelerates early descent and exploration; enforcing a minimum of 1e-6*span reduces chance of adaptation getting stuck on numerically negligible steps, improving robustness across varying dimensions.\\

\textbf{[Evaluation results]}
 \{"sphere\_d3\_i1": 0.699, "rosenbrock\_d5\_i2": 0.704, "rastrigin\_d10\_i5": 6.024, "ellipsoid\_d20\_i1": 1.343, "schaffers\_d40\_i5": 0.709\}

\end{level2box}
\caption{\textbf{Example of Delta Plan (Level 2).} This structured log details specific algorithmic modifications (Initialization, Polishing, Step-size), explicitly contrasting \texttt{OLD\_LOGIC} versus \texttt{NEW\_LOGIC} to expose the causal mechanism of improvement to the LLM.}
\label{fig:level2_example}
\end{figure}

\newpage
\section{Configuration of \deltaevolve}
\paragraph{Default Configuration.} Unless otherwise stated, all experiments use the default configuration summarized in \Cref{tab:system_config}.

\begin{table*}[h!]
\centering
\caption{System configuration and hyperparameters.}
\label{tab:system_config}
\begin{tabular}{l c l c}
\toprule
\textbf{Parameter} & \textbf{Value} & \textbf{Parameter} & \textbf{Value} \\
\midrule

\multicolumn{4}{l}{\textbf{LLM settings}} \\
\midrule
Primary model          & gpt-5-mini & Secondary model              & o3-mini \\
Temperature           & 0.7       & Top-p       & 0.95 \\
Max tokens        & 8192       & Timeout    & 600s \\

\midrule

\multicolumn{4}{l}{\textbf{Prompt}} \\
\midrule
Number of top plans              & 3   & Number of Diverse Plans     & 2 \\
Parent selection strategy & weighted & Elite selection ratio & 0.1   \\
Exploration ratio  &0.2 & Exploitation ratio &0.7 \\
\midrule

\multicolumn{4}{l}{\textbf{Database}} \\
\midrule
Population size              & 40   & Archive size     & 20 \\
Top-$k$ inspirations      & 3    &  Diverse inspirations      & 2 \\
Migration interval        & 10   & Migration Rate            & 0.1 \\
Number of islands         & 3 & Parent selection strategy & weighted \\

\midrule
\multicolumn{4}{l}{\textbf{Evolution}} \\
\midrule
Max Iterations                    & 100 & Random seed & 42 \\

\midrule

\multicolumn{4}{l}{\textbf{Evaluator}} \\
\midrule
Timeout                    & 300s & Max retries & 3 \\
Parallel evaluations & 4 & Cascade evaluation & False \\

\bottomrule
\end{tabular}
\end{table*}

\newpage

\section{Discovered Solutions}

For blackbox optimization, \deltaevolve discovery a new solution that achieves a score of 3.937, significantly surpassing the AlphaEvolve baseline of 2.6415. This gain stems from implementing a robust CMA-ES (Covariance Matrix Adaptation Evolution Strategy), which adapts the search distribution's geometry to match the problem's underlying landscape. By learning variable correlations and dynamically adjusting step sizes, the optimizer navigates ill-conditioned and non-separable functions far more efficiently than standard heuristics. Key enhancements include budget-aware hyperparameter scaling and a sophisticated boundary handling mechanism that preserves search direction, ensuring rapid convergence even under tight evaluation constraints.

\begin{lstlisting}[style=pyblue]
# EVOLVE-BLOCK-START
"""Baseline black-box optimizer for BBOB problems."""

from typing import Sequence, Tuple, Union
import numpy as np

def _get_bounds(problem, dimension: int) -> Tuple[np.ndarray, np.ndarray]:
    """Extract bounds from the problem or fall back to a symmetric box."""
    lower = getattr(problem, "lower_bounds", None)
    upper = getattr(problem, "upper_bounds", None)

    if lower is None or upper is None:
        lower = [-5.0] * dimension
        upper = [5.0] * dimension

    lower_arr = np.asarray(lower, dtype=float)
    upper_arr = np.asarray(upper, dtype=float)

    if lower_arr.shape[0] != dimension or upper_arr.shape[0] != dimension:
        lower_arr = np.full(dimension, float(lower_arr.flat[0]))
        upper_arr = np.full(dimension, float(upper_arr.flat[0]))

    return lower_arr, upper_arr

def _sample_uniform(rng: np.random.Generator, lower: np.ndarray, upper: np.ndarray) -> np.ndarray:
    """Uniform sample inside the box."""
    return lower + rng.random(size=lower.shape[0]) * (upper - lower)

def _clip_to_bounds(x: np.ndarray, lower: np.ndarray, upper: np.ndarray) -> np.ndarray:
    return np.clip(x, lower, upper)

def _to_params(x: np.ndarray) -> dict:
    """Convert vector to the dict format expected by bbob Problem.evaluate."""
    return {f"x{i}": float(v) for i, v in enumerate(x)}

def _evaluate_safe(problem, x: np.ndarray) -> float:
    """Evaluate the problem and guard against failures."""
    try:
        params = _to_params(x)
        value = problem.evaluate(params)
        value = float(value)
        if np.isnan(value) or np.isinf(value):
            return float("inf")
        return value
    except Exception:
        return float("inf")

def run_search(problem, budget: int = 1000, seed: int | None = None) -> Tuple[list[float], float, int]:
    """
    CMA-ES (Covariance Matrix Adaptation Evolution Strategy) optimizer.
    Adapts search distribution to efficiently find optimal solutions.
    Deterministic under `seed`.
    """
    rng = np.random.default_rng(seed)

    dimension = getattr(problem, "dimension", None)
    if dimension is None:
        lower_attr = getattr(problem, "lower_bounds", [])
        upper_attr = getattr(problem, "upper_bounds", [])
        dimension = len(lower_attr) or len(upper_attr) or 2

    lower, upper = _get_bounds(problem, dimension)

    # CMA-ES parameters
    # Initial sigma: A fraction of the search space width.
    # Adapt this based on budget and search space characteristics.
    average_range = np.mean((upper - lower) / 2.0)
    min_dim_range = np.min(upper - lower) # Added for a minimum step size

    if budget < 100 * dimension: # Heuristic for smaller budgets
        initial_sigma = average_range * 0.15 # Reduced multiplier for more controlled exploration
    else:
        initial_sigma = average_range * 0.4 # Reduced multiplier for more controlled exploration
    
    # Ensure a minimum initial sigma to prevent premature convergence on very small ranges
    initial_sigma = max(initial_sigma, min_dim_range * 0.05) # Reduced floor based on min dimension range
    
    # Population size (lambda). A common choice is 4 + floor(3 * log(dimension)).
    # Ensure population_size is at least 2 and allows for sufficient generations.
    base_population = int(4 + np.floor(3 * np.log(dimension)))
    population_size = max(2, base_population)
    if budget >= 10: # Only apply budget cap if budget is reasonable
        population_size = min(population_size, budget // 5) # More conservative cap to allow more generations
    if dimension > 20: # Cap for very high dimensions to manage computational cost
        population_size = min(population_size, 2 * dimension)
    
    # Initial mean: center of the search space
    mean = (lower + upper) / 2.0

    # Initialize CMA-ES state variables
    # C: Covariance matrix, initially identity matrix
    C = np.eye(dimension)
    # pc: Evolution path for C
    pc = np.zeros(dimension)
    # ps: Evolution path for sigma
    ps = np.zeros(dimension)
    # B: Eigenvectors of C (for coordinate system transformation)
    B = np.eye(dimension)
    # D: Square root of eigenvalues of C (scaling factors)
    D = np.ones(dimension)

    # Learning rates and weights
    # Number of parents (mu)
    mu = population_size // 2
    # Weights for recombination
    weights = np.log(mu + 0.5) - np.log(np.arange(1, mu + 1))
    weights = weights / np.sum(weights)
    # Variance effective selection mass
    mu_eff = np.sum(weights)**2 / np.sum(weights**2)

    # Adapt learning rates (constants from CMA-ES literature)
    # c_sigma: Learning rate for sigma
    c_sigma = (mu_eff + 2) / (dimension + mu_eff + 5)
    # d_sigma: Damping for sigma. Increased base value for slower step size reduction.
    d_sigma = 2.5 + 2 * max(0, np.sqrt((mu_eff - 1) / (dimension + 1)) - 1) + c_sigma
    # c_c: Learning rate for pc
    c_c = (4 + mu_eff / dimension) / (dimension + 4 + 2 * mu_eff / dimension)
    # c_1: Learning rate for C (rank-one update)
    # Learning rates (constants from CMA-ES literature)
    # c_sigma: Learning rate for sigma
    c_sigma = (mu_eff + 2) / (dimension + mu_eff + 5)
    # d_sigma: Damping for sigma. Increased base value for slower step size reduction.
    d_sigma = 2.5 + 2 * max(0, np.sqrt((mu_eff - 1) / (dimension + 1)) - 1) + c_sigma
    # c_c: Learning rate for pc
    c_c = (4 + mu_eff / dimension) / (dimension + 4 + 2 * mu_eff / dimension)
    # c_1: Learning rate for C (rank-one update)
    c_1 = 2 / ((dimension + 1.5)**2 + mu_eff)
    # c_mu: Learning rate for C (rank-mu update)
    c_mu = min(1 - c_1 - c_1 / mu, 2 * (mu_eff - 2 + 1 / mu_eff) / ((dimension + 2)**2 + mu_eff))

    evaluations_used = 0
    best_x = None
    best_value = float("inf")

    # The main CMA-ES loop
    while evaluations_used < budget:
        # Generate population
        population = []
        for _ in range(population_size):
            # Sample z from N(0, I)
            z = rng.normal(size=dimension)
            # Transform z to y in the search space
            y = B @ (D * z)
            # Add to mean and scale by sigma
            candidate_raw = mean + initial_sigma * y
            
            # Project candidates to bounds instead of clipping
            candidate = candidate_raw
            for i in range(dimension):
                if candidate[i] < lower[i]:
                    # Project towards mean along this dimension
                    candidate[i] = mean[i] + (lower[i] - mean[i]) * ((candidate_raw[i] - mean[i]) / (lower[i] - mean[i]))
                elif candidate[i] > upper[i]:
                    # Project towards mean along this dimension
                    candidate[i] = mean[i] + (upper[i] - mean[i]) * ((candidate_raw[i] - mean[i]) / (upper[i] - mean[i]))
            
            # Fallback clip to ensure strict bounds if projection logic is imperfect
            candidate = _clip_to_bounds(candidate, lower, upper)
            population.append(candidate)

        # Evaluate population
        fitnesses = []
        for candidate in population:
            if evaluations_used >= budget:
                break
            value = _evaluate_safe(problem, candidate)
            evaluations_used += 1
            fitnesses.append(value)

            if value < best_value:
                best_value = value
                best_x = candidate

        if evaluations_used >= budget:
            break

        # Sort population by fitness
        sorted_indices = np.argsort(fitnesses)
        sorted_population = [population[i] for i in sorted_indices]
        sorted_fitnesses = [fitnesses[i] for i in sorted_indices]

        # Select parents
        parents = sorted_population[:mu]

        # Update mean
        old_mean = mean
        mean_update = np.sum([weights[i] * (parents[i] - old_mean) for i in range(mu)], axis=0) / initial_sigma
        mean = old_mean + initial_sigma * mean_update

        # Update evolution paths
        ps_new = (1 - c_sigma) * ps + np.sqrt(c_sigma * (2 - c_sigma) * mu_eff) * (B @ mean_update)
        ps = ps_new

        hsig = np.linalg.norm(ps) / np.sqrt(1 - (1 - c_sigma)**(2 * evaluations_used / population_size)) / np.sqrt(dimension) < (1.4 + 2 / (dimension + 1))
        
        pc_new = (1 - c_c) * pc + hsig * np.sqrt(c_c * (2 - c_c) * mu_eff) * mean_update
        pc = pc_new

        # Update covariance matrix C
        C_old = C
        C_rank_one = np.outer(pc, pc) if dimension > 1 else np.array([[pc[0] * pc[0]]])
        
        C_rank_mu = np.sum([weights[i] * np.outer((parents[i] - old_mean) / initial_sigma, (parents[i] - old_mean) / initial_sigma) for i in range(mu)], axis=0)

        C = (1 - c_1 - c_mu) * C_old + c_1 * C_rank_one + c_mu * C_rank_mu
        
        # Ensure C is symmetric and positive definite (numerical stability)
        C = (C + C.T) / 2.0 + 1e-10 * np.eye(dimension) # Add small diagonal to ensure positive definiteness

        # Update sigma
        initial_sigma = initial_sigma * np.exp((c_sigma / d_sigma) * (np.linalg.norm(ps) / np.sqrt(dimension) - 1))

        # Eigendecomposition of C for B and D
        # Re-evaluate B and D periodically to save computation.
        # This is typically done if the number of evaluations is a multiple of some factor,
        # or every few generations. For simplicity and robustness, we do it every generation.
        # Note: np.linalg.eigh returns eigenvalues in ascending order.
        D_squared, B = np.linalg.eigh(C)
        D = np.sqrt(np.maximum(1e-10, D_squared)) # Ensure strictly positive eigenvalues for numerical stability

    # Fallback if everything failed/returned inf
    if best_x is None or not np.isfinite(best_value):
        best_x = _clip_to_bounds((lower + upper) / 2.0, lower, upper)
        best_value = _evaluate_safe(problem, best_x)
        if not np.isfinite(best_value):
            best_value = float("inf")

    return best_x.tolist(), float(best_value), evaluations_used

# EVOLVE-BLOCK-END

def run_search_entry(problem, budget: int = 1000, seed: int | None = None):
    """
    Thin wrapper kept outside the evolve block in case the block is replaced.
    """
    return run_search(problem, budget=budget, seed=seed)

if __name__ == "__main__":
    # Smoke test with a trivial sphere if optunahub is available
    try:
        import optunahub

        bbob = optunahub.load_module("benchmarks/bbob")
        test_problem = bbob.Problem(function_id=1, dimension=3, instance_id=1)
        x, value, used = run_search(test_problem, budget=100, seed=0)
        print(f"Best value {value:.4e} after {used} evals at x={x}")
    except Exception as exc:
        print(f"Smoke test skipped ({exc})")

\end{lstlisting}

%% file: reference.bib
@article{ellis2023dreamcoder,
  title={Dreamcoder: growing generalizable, interpretable knowledge with wake--sleep bayesian program learning},
  author={Ellis, Kevin and Wong, Lionel and Nye, Maxwell and Sable-Meyer, Mathias and Cary, Luc and Anaya Pozo, Lore and Hewitt, Luke and Solar-Lezama, Armando and Tenenbaum, Joshua B},
  journal={Philosophical Transactions of the Royal Society A},
  volume={381},
  number={2251},
  pages={20220050},
  year={2023},
  publisher={The Royal Society}
}

@article{shi2022compositional,
  title={Compositional generalization and decomposition in neural program synthesis},
  author={Shi, Kensen and Hong, Joey and Zaheer, Manzil and Yin, Pengcheng and Sutton, Charles},
  journal={arXiv preprint arXiv:2204.03758},
  year={2022}
}

@article{reed2015neural,
  title={Neural programmer-interpreters},
  author={Reed, Scott and De Freitas, Nando},
  journal={arXiv preprint arXiv:1511.06279},
  year={2015}
}

@article{press2025algotune,
  title={AlgoTune: Can Language Models Speed Up General-Purpose Numerical Programs?},
  author={Press, Ori and Amos, Brandon and Zhao, Haoyu and Wu, Yikai and Ainsworth, Samuel K and Krupke, Dominik and Kidger, Patrick and Sajed, Touqir and Stellato, Bartolomeo and Park, Jisun and others},
  journal={arXiv preprint arXiv:2507.15887},
  year={2025}
}

@techreport{bbob2019,
    author = {Finck, Steffen and Hansen, Nikolaus and Ros, Raymond and Auger, Anne},
    title = {Real-Parameter Black-Box Optimization Benchmarking 2009: Noiseless Functions Definitions},
    institution = {INRIA},
    year = {2009},
    number = {RR-6829},
    note = {Updated version as of February 2019},
    url = {https://inria.hal.science/inria-00362633v2/document}
}

@article{shojaee2025llm,
  title={Llm-srbench: A new benchmark for scientific equation discovery with large language models},
  author={Shojaee, Parshin and Nguyen, Ngoc-Hieu and Meidani, Kazem and Farimani, Amir Barati and Doan, Khoa D and Reddy, Chandan K},
  journal={arXiv preprint arXiv:2504.10415},
  year={2025}
}

@article{shi2025no,
  title={No Loss, No Gain: Gated Refinement and Adaptive Compression for Prompt Optimization},
  author={Shi, Wenhang and Chen, Yiren and Bian, Shuqing and Zhang, Xinyi and Tang, Kai and Hu, Pengfei and Zhao, Zhe and Lu, Wei and Du, Xiaoyong},
  journal={arXiv preprint arXiv:2509.23387},
  year={2025}
}

@article{fei2025efficient,
  title={Efficient Prompt Compression with Evaluator Heads for Long-Context Transformer Inference},
  author={Fei, Weizhi and Niu, Xueyan and Xie, Guoqing and Liu, Yingqing and Bai, Bo and Han, Wei},
  journal={arXiv preprint arXiv:2501.12959},
  year={2025}
}

@article{zhang2024long,
  title={Long context compression with activation beacon},
  author={Zhang, Peitian and Liu, Zheng and Xiao, Shitao and Shao, Ninglu and Ye, Qiwei and Dou, Zhicheng},
  journal={arXiv preprint arXiv:2401.03462},
  year={2024}
}

@article{wang2025m+,
  title={M+: Extending MemoryLLM with Scalable Long-Term Memory},
  author={Wang, Yu and Krotov, Dmitry and Hu, Yuanzhe and Gao, Yifan and Zhou, Wangchunshu and McAuley, Julian and Gutfreund, Dan and Feris, Rogerio and He, Zexue},
  journal={arXiv preprint arXiv:2502.00592},
  year={2025}
}

@article{wang2024memoryllm,
  title={Memoryllm: Towards self-updatable large language models},
  author={Wang, Yu and Gao, Yifan and Chen, Xiusi and Jiang, Haoming and Li, Shiyang and Yang, Jingfeng and Yin, Qingyu and Li, Zheng and Li, Xian and Yin, Bing and others},
  journal={arXiv preprint arXiv:2402.04624},
  year={2024}
}

@online{anthropic2025context,
  author       = {Anthropic},
  title        = {Effective Context Engineering for AI Agents},
  year         = {2025},
  month        = {Sep},
  day          = {29},
  url          = {https://www.anthropic.com/engineering/effective-context-engineering-for-ai-agents}
}

@article{ramnath2025systematic,
  title={A systematic survey of automatic prompt optimization techniques},
  author={Ramnath, Kiran and Zhou, Kang and Guan, Sheng and Mishra, Soumya Smruti and Qi, Xuan and Shen, Zhengyuan and Wang, Shuai and Woo, Sangmin and Jeoung, Sullam and Wang, Yawei and others},
  journal={arXiv preprint arXiv:2502.16923},
  year={2025}
}

@article{yuksekgonul2026learning,
  title={Learning to Discover at Test Time},
  author={Yuksekgonul, Mert and Koceja, Daniel and Li, Xinhao and Bianchi, Federico and McCaleb, Jed and Wang, Xiaolong and Kautz, Jan and Choi, Yejin and Zou, James and Guestrin, Carlos and others},
  journal={arXiv preprint arXiv:2601.16175},
  year={2026}
}

@article{hansen2016cma,
  title={The CMA evolution strategy: A tutorial},
  author={Hansen, Nikolaus},
  journal={arXiv preprint arXiv:1604.00772},
  year={2016}
}

@article{mankowitz2023faster,
  title={Faster sorting algorithms discovered using deep reinforcement learning},
  author={Mankowitz, Daniel J and Michi, Andrea and Zhernov, Anton and Gelmi, Marco and Selvi, Marco and Paduraru, Cosmin and Leurent, Edouard and Iqbal, Shariq and Lespiau, Jean-Baptiste and Ahern, Alex and others},
  journal={Nature},
  volume={618},
  number={7964},
  pages={257--263},
  year={2023},
  publisher={Nature Publishing Group UK London}
}

@inproceedings{real2020automl,
  title={Automl-zero: Evolving machine learning algorithms from scratch},
  author={Real, Esteban and Liang, Chen and So, David and Le, Quoc},
  booktitle={International conference on machine learning},
  pages={8007--8019},
  year={2020},
  organization={PMLR}
}

@inproceedings{averly2025liddia,
  title={Liddia: Language-based intelligent drug discovery agent},
  author={Averly, Reza and Baker, Frazier N and Watson, Ian A and Ning, Xia},
  booktitle={Proceedings of the 2025 Conference on Empirical Methods in Natural Language Processing},
  pages={12015--12039},
  year={2025}
}

@article{li2025codepde,
  title={CodePDE: An Inference Framework for LLM-driven PDE Solver Generation},
  author={Li, Shanda and Marwah, Tanya and Shen, Junhong and Sun, Weiwei and Risteski, Andrej and Yang, Yiming and Talwalkar, Ameet},
  journal={arXiv preprint arXiv:2505.08783},
  year={2025}
}

@article{bansal2025let,
  title={Let's (not) just put things in Context: Test-Time Training for Long-Context LLMs},
  author={Bansal, Rachit and Zhang, Aston and Tiwari, Rishabh and Madaan, Lovish and Duvvuri, Sai Surya and Khatri, Devvrit and Brandfonbrener, David and Alvarez-Melis, David and Bhargava, Prajjwal and Kale, Mihir Sanjay and others},
  journal={arXiv preprint arXiv:2512.13898},
  year={2025}
}

@article{georgiev2025mathematical,
  title={Mathematical exploration and discovery at scale},
  author={Georgiev, Bogdan and G{\'o}mez-Serrano, Javier and Tao, Terence and Wagner, Adam Zsolt},
  journal={arXiv preprint arXiv:2511.02864},
  year={2025}
}

@article{romera2024mathematical,
  title={Mathematical discoveries from program search with large language models},
  author={Romera-Paredes, Bernardino and Barekatain, Mohammadamin and Novikov, Alexander and Balog, Matej and Kumar, M Pawan and Dupont, Emilien and Ruiz, Francisco JR and Ellenberg, Jordan S and Wang, Pengming and Fawzi, Omar and others},
  journal={Nature},
  volume={625},
  number={7995},
  pages={468--475},
  year={2024},
  publisher={Nature Publishing Group UK London}
}

@article{novikov2025alphaevolve,
  title={AlphaEvolve: A coding agent for scientific and algorithmic discovery},
  author={Novikov, Alexander and V{\~u}, Ng{\^a}n and Eisenberger, Marvin and Dupont, Emilien and Huang, Po-Sen and Wagner, Adam Zsolt and Shirobokov, Sergey and Kozlovskii, Borislav and Ruiz, Francisco JR and Mehrabian, Abbas and others},
  journal={arXiv preprint arXiv:2506.13131},
  year={2025}
}

@article{hubert2025olympiad,
  title={Olympiad-level formal mathematical reasoning with reinforcement learning},
  author={Hubert, Thomas and Mehta, Rishi and Sartran, Laurent and Horv{\'a}th, Mikl{\'o}s Z and {\v{Z}}u{\v{z}}i{\'c}, Goran and Wieser, Eric and Huang, Aja and Schrittwieser, Julian and Schroecker, Yannick and Masoom, Hussain and others},
  journal={Nature},
  pages={1--3},
  year={2025},
  publisher={Nature Publishing Group UK London}
}

@article{trinh2024solving,
  title={Solving olympiad geometry without human demonstrations},
  author={Trinh, Trieu H and Wu, Yuhuai and Le, Quoc V and He, He and Luong, Thang},
  journal={Nature},
  volume={625},
  number={7995},
  pages={476--482},
  year={2024},
  publisher={Nature Publishing Group UK London}
}

@software{openevolve,
  title = {OpenEvolve: an open-source evolutionary coding agent},
  author = {Asankhaya Sharma},
  year = {2025},
  publisher = {GitHub},
  url = {https://github.com/algorithmicsuperintelligence/openevolve}
}

@article{lange2025shinkaevolve,
  title={Shinkaevolve: Towards open-ended and sample-efficient program evolution},
  author={Lange, Robert Tjarko and Imajuku, Yuki and Cetin, Edoardo},
  journal={arXiv preprint arXiv:2509.19349},
  year={2025}
}

@article{assumpccao2025codeevolve,
  title={CodeEvolve: An open source evolutionary coding agent for algorithm discovery and optimization},
  author={Assump{\c{c}}{\~a}o, Henrique and Ferreira, Diego and Campos, Leandro and Murai, Fabricio},
  journal={arXiv preprint arXiv:2510.14150},
  year={2025}
}

@article{khrulkov2025gigaevo,
  title={GigaEvo: An Open Source Optimization Framework Powered By LLMs And Evolution Algorithms},
  author={Khrulkov, Valentin and Galichin, Andrey and Bashkirov, Denis and Vinichenko, Dmitry and Travkin, Oleg and Alferov, Roman and Kuznetsov, Andrey and Oseledets, Ivan},
  journal={arXiv preprint arXiv:2511.17592},
  year={2025}
}

@article{wang2025thetaevolve,
  title={ThetaEvolve: Test-time Learning on Open Problems},
  author={Wang, Yiping and Su, Shao-Rong and Zeng, Zhiyuan and Xu, Eva and Ren, Liliang and Yang, Xinyu and Huang, Zeyi and He, Xuehai and Ma, Luyao and Peng, Baolin and others},
  journal={arXiv preprint arXiv:2511.23473},
  year={2025}
}
